\documentclass[a4paper,fleqn]{cas-dc}
\usepackage{subfigure}
\usepackage{algorithm}
\usepackage{algpseudocode}
\algnewcommand{\Input}{\item[\textbf{Input:}]}
\algnewcommand{\Output}{\item[\textbf{Output:}]}
\usepackage{xcolor}

\usepackage[numbers]{natbib}
\def\tsc#1{\csdef{#1}{\textsc{\lowercase{#1}}\xspace}}
\tsc{WGM}
\tsc{QE}
\begin{document}
\let\WriteBookmarks\relax
\def\floatpagepagefraction{1}
\def\textpagefraction{.001}
\let\printorcid\relax  

\shorttitle{}    

\shortauthors{}  

\title [mode = title]{DNMDR: Dynamic Networks and Multi-view Drug Representations for Safe Medication Recommendation}  




\author[1]{Guanlin Liu}%[<options>]
%\ead{liuguanlin0818@163.com}

\author[1,2]{Xiaomei Yu}
\cormark[1]
\ead{yxm0708@126.com}
\author[1]{Zihao Liu}
\author[1]{Xue Li}
\author[1]{Xingxu Fan}
\author[1,2,3]{Xiangwei Zheng}

\cortext[1]{Corresponding author.}

\affiliation[1]{organization={School of Information Science and Engineering, Shandong Normal University},
            city={Jinan},
%          citysep={}, % Uncomment if no comma needed between city and postcode
            postcode={250358}, 
            country={China}}
            
\affiliation[2]{organization={Shandong Provincial Key Laboratory for Distributed Computer Software Novel Technology},
            city={Jinan},
            postcode={250358}, 
            country={China}}

\affiliation[3]{organization={State Key Laboratory of High-end Server and Storage Technology},
            city={Jinan},
            postcode={250300}, 
            country={China}}




\begin{abstract}
Medication Recommendation (MR) is a promising research topic which booms diverse applications in the healthcare and clinical domains. However, existing methods mainly rely on sequential modeling and static graphs for representation learning, which ignore the dynamic correlations in diverse medical events of a patient’s temporal visits, leading to insufficient global structural exploration on nodes. Additionally, mitigating drug-drug interactions (DDIs) is another issue determining the utility of the MR systems. To address the challenges mentioned above, this paper proposes a novel MR method with the integration of dynamic networks and multi-view drug representations (DNMDR). Specifically, weighted snapshot sequences for dynamic heterogeneous networks are constructed based on discrete visits in temporal EHRs, and all the dynamic networks are jointly trained to gain both structural correlations in diverse medical events and temporal dependency in historical health conditions, for achieving comprehensive patient representations with both semantic features and structural relationships. Moreover, combining the drug co-occurrences and adverse drug-drug interactions (DDIs) in internal view of drug molecule structure and interactive view of drug pairs, the safe drug representations are available to obtain high-quality medication combination recommendation. Finally, extensive experiments on real world datasets are conducted for performance evaluation, and the experimental results demonstrate that the proposed DNMDR method outperforms the state-of-the-art baseline models with a large margin on various metrics such as PRAUC, Jaccard, DDI rates and so on. 
\end{abstract}


%%%%\begin{highlights}
%%%%\item A novel recommendation model named DNMDR is proposed for accuracy and safe drug combination recommendation. 

%%%%\item  Weighted dynamic networks are leveraged to  capture the evolution of a patient’s historical conditions in visits.

%%%%\item The Multi-view drug  co-occurrence and DDIs  are learned with GATs to generate comprehensive drug representations.

%%%%\item The extensive experiments are conducted and the experimental results reveal the superior performance of DNMDR.
%%%%\end{highlights}

%\nocite{*}

% Keywords

\maketitle

% Main text
\section{Introduction}\label{sec1}
With the booming of information collection on electronic health records (EHRs), medical data have been widely accumulated and the quality of EHR data have been constantly improved, which promote the research progress on medication recommendation (MR). An MR system learns historical medical events and prescription data in EHRs, and offers drug combinations which might be significant for a physician. On the one hand, the MR system supports the clinical physicians with accurate medication prescriptions for patients in complex health conditions efficiently. On the other hand, a well trained MR system plays crucial role in automatically reviewing prescriptions on safety, and ensures that the prescribed medications are consistent with the patients’ diagnostic details and correspond to the patient’s physical concerns.

Recent years have witnessed the opportunity of improving efficiency and quality of drug recommendation. In the patient representation component, recurrent neural network (RNN) based variants are the dominant longitudinal models \cite{li2024transformer} \cite{10614809} \cite{9995543}, which demonstrate outstanding advantages on capturing temporality and dependency in patients’ historical visits, and boost the semantic representation ability on patients. In the drug combination representation component, both semantic information and structural relationships are concerned in impressive studies such as SafeDrug \cite{yang2021safedrug} and GAMENet \cite{shang2019gamenet}, and the Graph Convolutional Network (GCN) is beneficial to integral drug representations learning. Combining the representations of both patients and drugs with a multi-label classifier, a list of drugs is predicted for recommendation\cite{yu2024aka}. Though considerable research fruits gained, there are still challenging problems worth further investigation to promote the performance of MR in practical healthcare scenarios\cite{nguyen2016deepr}.

First, collecting thorough health records of each patient are extremely difficult\cite{gong2021smr}, and it is challenging to offer personalized recommendation results for patients with sparse visits in EHRs. On the one hand, though considerable improvements have gained on learning semantic information with sequential EHR data, it is due to the huge effort dedicated to handle the explicitly irregular EHRs, rather than to consider the potential correlations in clinical events. On the other hand, it is the intricate dependencies in diverse medical events that could benefit the patient's current visit in diagnosis supporting and prescription making. Moreover, the potential structural information is ignored in representation learning on patients with sparse and incomplete EHRs. Hence, the imperfect patient representations hamper the prediction performance in drug recommendation, even though external knowledge support is provided in the recommendation process.

Second, the evolution of a patient’s diseases is a dynamic and constantly changing process, which should be considered in EHR data modeling. In fact, the quality of clinical data is based on variable factors, such as clinical measurements, medical examinations and professional expertise, which are subjective and may be influenced by the preferences of physicians. For example, the absence of a diagnosis may mean that a person has not yet been diagnosed with a disease, which does not always mean the absence of a disease on the person. Additionally, most of the studies capture the sequential nature in the patients’ visits, while they disregard the irregular time factors in clinical records. Since the time intervals between clinical events carry vital information about the progression of a patient’s health conditions, data irregularity has already been pointed out as a major challenge in modeling temporal data. Furthermore, a minority of existing studies leverage structural information in constructed static heterogeneous graphs. They learn  fixed connections in medical event nodes, while disregard the variations of clinical characters in representation learning for patients and drugs. Therefore, the dynamicity and possible evolution in patients’ sequential visits worth further exploration to promote prediction performance.

Third, with the constant emergence of new drugs and innovative usages on present drugs, the existing medical knowledge is inadequate to reach accurate and safe medication representations. On the one hand, the prior knowledge from experienced clinicians should be investigated to ensure drug co-occurrences and to prevent adverse DDIs in innovative usage. On the other hand, the potential DDIs unknown at present should be further explored to guarantee  medication safety in personalized prescriptions. Therefore, the exploration on drug representations from diverse levels and multiple perspectives are necessary based on both prescribed experience in EHR data and known DDIs in external database.

To alleviate the limitations mentioned above, this paper proposes a novel MR model which integrates dynamic networks with multi-view drug representations (DNMDR). Specifically, considering irregular time intervals between a patient’s visits, the dynamic networks offer a comprehensive representation on both temporal dependencies of historical visits and dynamical correlations of clinical events, with the guidance of conditional probabilities to extract implicit structural information in EHR data. Additionally, both EHR and DDI graphs are leveraged to investigate prior knowledge in experienced prescriptions and known DDIs in medical databases, respectively. Moreover, incorporating molecular-level function from a structural perspective and patient-drug interaction information from a semantic perspective, the integral representation learning on drugs enables accurate and safe medication predictions with a multi-label classifier.

The main contributions of the paper are summarized as follows:

\begin{itemize}
\item Based on the weighted dynamic networks and a multi-view drug graph, a novel recommendation model is proposed for accurate and safe drug combination recommendation. Integrating integral patient representations learned in dynamical clinical events and historical EHR sequences, with multi-level drug representation from semantic and structural perspectives, the proposed DNMDR model facilitates accurate and safe drug recommendation.

\item Based on implicit structural information extracted with the guidance of conditional probabilities, the weighted dynamic networks are constructed to settle the data irregularity issue in sparse and incomplete EHR data. Moreover, RNNs are employed to effectively capture the evolution of a patient’s historical conditions in visits with irregular time intervals. As a result, the weighted dynamic networks model the patient representations accurately for downstream tasks.

\item Based on explored clinical experience and medical knowledge from multiple perspectives, the drug co-occurrence and adverse DDIs in diverse levels are available to generate comprehensive drug representations. Integrating drug embeddings encoded with corresponding drug atoms, chemical bones and molecular structures, the introduced multi-view drug graph enables thorough medical knowledge learning on prior experience, known DDI knowledge and potential DDI predictor, and further facilitates drug recommendation eventually.

\item The extensive experiments are conducted on real-life EHR dataset, and the experimental results reveal the superior performance of DNMDR, compared with the existing state-of-the-art baseline models on different evaluation metrics, such as DDI rates, Jaccard similarity and F1 score.
\end{itemize}

The rest of this paper is organized as follows. In section \ref{sec2}, a summary of the related work is reviewed. Section \ref{sec3} presents the preliminary knowledge. And the DNMDR model is detailed in Section \ref{sec4}. In Section \ref{sec5}, the experiments are conducted on real-world EHR datasets and the results are analyzed. Finally, the conclusion is drawn and the future work is pointed out in Section \ref{sec6}.

\section{Related Work}\label{sec2}
Equipped with two key components, the MR systems predict a set of drugs corresponding to the patient’s diagnostic concerns, which reduces the time of prescription making. With prevailing studies achieved, the related work in this paper focuses on two lines: research on patient representation and drug representation.

\subsection{Patient representation}\label{subsec21}
In prevailing studies on patient representations, a majority of longitudinal methods are employed to encode EHR sequential data with RNN-based variants\cite{yang2021safedrug,shang2019gamenet,yu2024aka}. Choi et al.\cite{choi2016retain} put forward a two-level attention mechanism on RNN models to learn temporal characters from multiple admissions, which captures semantic features that affect the patient’s health conditions, with no deliberate consideration on long-range dependency in historical visits. To leverage the large number of EHRs with a single visit for each patient, G-BERT\cite{shang2019pre} is put forward which combines Graph Neural Networks (GNNs) with Bidirectional Encoder Representations from Transformers (BERT) for medical code representation and further medication recommendation. With internal hierarchical structures of medical codes fed, the transformer-based encoders are then fine-tuned for downstream predictive tasks. Though semantic features are fully explored on longitudinal EHRs with multiple visits, the structural correlations in different clinical events are disregarded in G-BERT, which hinders the boosting of recommendation performance. To generate an adequate representation on input temporal data, COGNet are employed with two Transformers to mine the relationships in patients’ historical medical events and current physics status. However, these methods disregard structural information and varying dependencies in the disease evolution and medical development, thereby furnishing no solution for data irregularity issue in diverse clinical events.

Generally, the EHR data consist of diverse temporal sequences and multilevel structure information, including data with flat structure in each sequence and hierarchical structure between sequences, which implicitly reflect a physician’s decision procession. To deeply investigate the internal hierarchical structure of medical codes, GNNs have emerged as a powerful tool to model EHR data as graphs. Besides G-BERT \cite{shang2019pre}, MendMKG \cite{MendMKG} conducts a self-supervised learning strategy and pre-trains a graph attention network (GAT) for node embeddings, thereby complete the medical knowledge graph and reconstruct the missing edges in EHRs. Zhang et. al. \cite{zhang2023enhancing} employed a heterogeneous information network (HIN) to represent multi-modal relationships among various EHR nodes and proposed a graph representation learning method, in which a bi-channel heterogeneous local structural encoder is developed to extract diverse structural information in EHRs. On consideration that there is not always complete structure information in EHR data, the Graph Convolutional Transformer is addressed which leverages data statistics to guide the structure learning process. All in all, the graph-based representation learning methods focus on complete semantic features and intricate structure relationships in EHRs, and contribute to the solution of data irregularity issue. Nevertheless, the time internal irregularity is still a challenging issue which is worth further exploration.

Benefiting from the existing studies, the dynamic networks arise naturally to model nodes, attributes and edges which describes the evolving health status with time. The last few years have witnessed a surge of researches on dynamic networks including relation prediction\cite{pareja2020evolvegcn,gao2022novel}, node classification\cite{kazemi2020representation} and recommendation systems\cite{guo2024dynamic}. In this paper, to modelthe patients’ continuously evolving health conditions in temporal EHR data and to concern the intricate structural relationships as well as rich dependency in clinical events of each visit, we develop EHR dynamic networks to perform graph representation learning on patients. Based on weighted structure learning with conditional probability, the paper inclines to settle the data irregularity and time irregularity issues for accurate and safe MR tasks.

\subsection{Drug representation}\label{subsec22}
With suitable data structure in non-Euclidean space, the medical data are generally modeled with graph convolution neural networks (GCNs) in prevailing studies. After each node is initialized, GCN updates the nodes by neighbor iterative aggregation, and finally generates the informative latent feature representations for each node \cite{wu2022conditional, wang2021multi, wang2018personalized, choi2020learning, pareja2020evolvegcn}. In this research direction, the work of Duvenaud et al.\cite{duvenaud2015convolutional} captures deep semantic features of medication fingerprints with GCN. And Huang et al.\cite{huang2020deeppurpose} constructed a medication molecule graph andwhich captures medical features in drug pairs, based on GCN\cite{huang2020caster}.

Avoiding fatal DDIs is among the prominent challenges in MR tasks, and the graph-based learning methods have aroused widespread interest which are priority options for DDI controller. Zitnik et al.\cite{zitnik2018modeling} leveraged GCNs to explore the side effects in drug pairs with a two-layer DDI graph. SafeDrug is equipped with a global message passing neural network and a local binary learning module to comprehensively encode the structural connectivity and functional properties of drug molecules. GAMENet applies a two-layer GCN on each graph to capture global semantic information on drug co-occurrences and drug side effects respectively. In general, the GNN models employed in DDI prediction component are inclined to obtain medication representations with integral drug features. However, GNNs encounter several limitations in drug representation learning. On the one hand, GNNs are intensive in computation, and their complex architecture often demands a substantial amount of data for effective training. On the other hand, the memory space for possible drug combinations is vast, and only a small fraction of potential drug pairs can be tested experimentally. Therefore, with limited data and complex models, there is a high risk of overfitting, which makes it difficult for GNNs to make accurate predictions, especially for untested combinations.

These limitations arouse active exploration on various techniques to enhance the learning ability of GNNs. In this research direction, ALGNet combines a light graph convolutional network (LGCN) with a reinforcement memory network to encode the patients’ medical records and DDI knowledge for safe MR tasks. Graph Attention Networks (GATs) measure each neighbor node with attention weight and capture important structural information with relative scores in graph structures. In GraphCare \cite{jiang2023graphcare}, a bi-attention augmented GNN is proposed with both visit-level and node-level attentions. Along with edge weights benefiting from personalized drug knowledge graph, GraphCare enhances healthcare prediction in scenarios with limited medical data. Superior to GCN, the GAT models calculate the local network of nodes with promoted calculation efficiency and reduced memory usage, which is beneficial to generate graph representations with enhanced learning ability. Moreover, GAT assigns different weights to various neighbor nodes according to the importance of the current node, which benefits the GAT model in dealing with structural learning problems, and make it suitable to analyze EHR data in the contexts.

With attention mechanism incorporated into the GCN, GATs make the propagation step more intuitive. Different from GCN, the GAT model is better in dealing with the structural problems. Therefore, a multi-view graph representation learning method is proposed in this paper. Integrating structural information of atoms, chemical bones and drug molecules, the nodes in drug graphs are encoded with GATs,  including drug co-occurrences and DDIs, which facilitates the final medication representation.

\section{Problem Formulation}\label{sec3}
Most of the existing studies focus on static graphs, where the structure of vertices is fixed. However, in practical clinical scenarios, the EHR data are dynamic and evolving over time.

\subsection{Notation definition}\label{subsec31}
\subsubsection{A dynamic network for a patient’s EHR data}\label{subsec311}
\begin{figure}
\centering
\includegraphics[width=.48\textwidth]{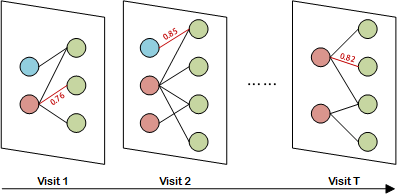}
\caption{A dynamic network for a single patient.}\label{fig1}
\end{figure}

The sequential visits for patients are recorded as a set of dynamic networks $R=\{G_{1},G_{2},\cdots,G_{N}\}$, where $N$ is the total number of patients in the EHRs, and each patient consists of multiple medical events in diverse medical codes. In this paper, we take a single patient with multiple visits as an example. Figure \ref{fig1} illustrates the dynamic structure of a patient’s EHR data in continuous visits. The dynamic network of the patient with diverse medical records consists of a series of snapshots $G_{i}=\{G_{i}^{1},\cdots,G_{i}^{T}\}$, where $G_{i}^{t}=(V_{t},E_{t},W_{t}) (1\leq t\leq T, 1\leq i\leq N)$ indicates an undirected graph with diverse nodes $V_{t}$ connecting weighted edges $E_{t}$ at $t$-th visit of the patient $i$. In order to simplify the notation, we remove the subscript $i$ of $G_{i}^{t}$ with no ambiguity. Here, each node is a temporal medical event expressed as medical codes $V_{t}=[d_{t},p_{t},m_{t}]$, where $d_{t}\in\{0,1\}^{|D|}$, $p_{t}\in\{0,1\}^{|P|}$ and $m_{t}\in\{0,1\}^{|M|}$ are the multi-hot vectors, and $D$, $P$, $M$ represents the set of diagnosis codes, procedure codes, medication codes, respectively. Additively,  each edge $e_{ij}^t\in E_{t}$ is associated with a adjacency matrix $\hat{A}_{t}\in \mathbb{R}^{|V_{t}|\times|V_{t}|}$, indicating the probability of hidden structural information between the nodes of diverse medical codes. Moreover, an weight $w_{ij}^{t} := \tilde{w}_{ij}^{t}({e_{ij}})({w_{ij}^{t}}\in{W_{t}})$ denotes the probability of positive or negative relationships between the two neighboring nodes. Here, a mapping function $\tilde{w}^t(\cdot) : E_{t} \mapsto \mathbb{R}$ is utilized to manifest the mapping from the edges to the real values.

In general, a patient’s dynamic network $G^{i}$($i\in$ $\{{1},{2},\cdots,{T}\}$) represents the semantic features and structural relationships of medical codes in the visit sequence, where $T$ is the total number of visits for a patient. 

\begin{table}
\caption{Notations used in this paper}\label{tbl1}
\begin{tabular*}{\tblwidth}{@{}C|L@{}}
\toprule
Notation & Description \\ 
\midrule
$R$ & a set of dynamic networks with EHRs \\ 
$G_i$ & a dynamic network for a single patient \\
$\hat{A}_{t}$ & the hidden structural information of $t$-th snapshot \\
$D,P,M$ & Diagnoses, Procedure and Medication set \\ 
$d,p,m$ & Diagnoses, Procedure and Medication code \\ 
$G_{ehr}, G_{ddi}$ & EHR graph and DDI graph \\ 
$T$ & the number of visits \\ 
\bottomrule
\end{tabular*}
\end{table}

\subsubsection{EHR \& DDI Graphs}\label{subsec312}
The EHR graph $G_{ehr}=\left\{M,A_{ehr}\right\}$ is constructed to illustrate the co-occurred drug pairs in a prescription of a patient’s single visit. Similarly, the DDI graph $G_{ddi}=\left\{M,A_{ddi}\right\}$ is constructed to clarify the known adverse DDIs recorded in the drug database\cite{tatonetti2012data}, where $A_{ehr},A_{ddi}\in \mathbb{R}^{|M|\times|M|}$ are both adjacency matrix. Here, $A_{ehr}\left[i,j\right]=1$ denotes the co-prescribed relationships of drug $i$ and drug $j$ by a physician in the EHRs and $A_{ddi}\left[i,j\right]=1$ indicates the adverse interaction between drug $i$ and drug $j$ that should be avoided. Following the  previous studies, we consider the known DDIs in pairwise drugs according to the public drug databases.

The notation used in this paper are summarized in Table \ref{tbl1}.

\subsection{Problem statement}
\label{subsec32}
Given the patients’ historical visits from the first to the $(t-1)$ admissions with diagnosis codes $d_{1:t-1}$, procedure codes $p_{1:t-1}$ and medication codes $m_{1:t-1}$, for a patient’s current diagnosis codes $d_{t}$, procedure codes $p_{t}$, the objective of this paper is to learn a function $f(\cdot)$ for medication recommendation, i.e., to predict accurate and safe medication combination $\hat{y}_{t}$ to the patient with a multi-label classifier. The process is formalized as follows:
\begin{equation}\hat{y}_{t}=f(d_{1:t},p_{1:t},m_{1:t-1})\label{eq:equation1}\end{equation}
where $\hat{y_t}\in\{0,1\}^{|M|}$ is the multi-label output.

\section{The DNMDR Model}\label{sec4}
To settle the inadequate structural learning issue on patients and drugs, while avoid the data irregularity and learn time dependence, a novel drug recommendation model, named DNMDR, is proposed in this section. Firstly, the architecture of DNMDR is demonstrated. Subsequently, the key components are detailed with principle, composition and contributions. Finally, the main algorithm for an epoch training is outlined.

\subsection{Architecture}
\label{subsec41}
\begin{figure*}
\centering
\includegraphics[scale=0.43]{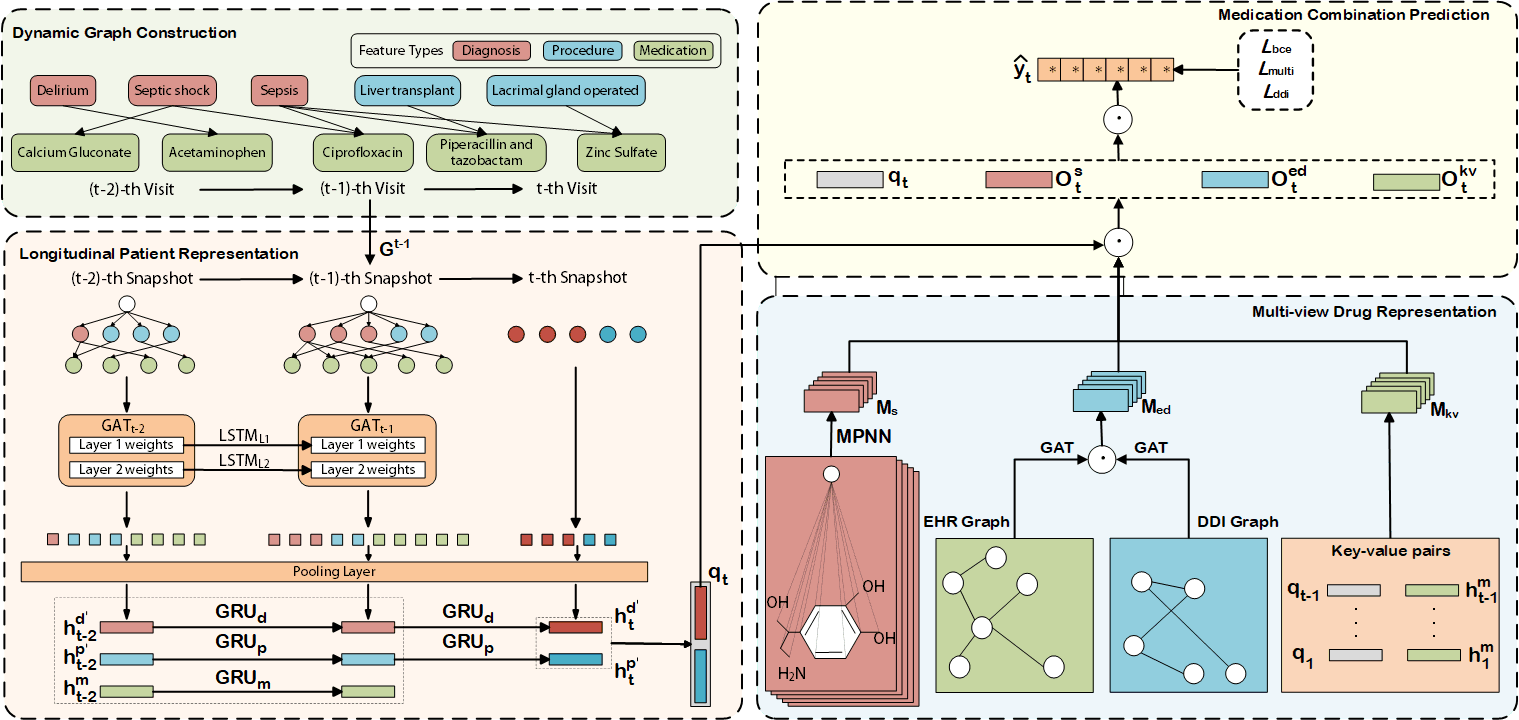}
\caption{The architecture of DNMDR}\label{fig2}
\end{figure*}

The DNMDR model consists of four components: dynamic graph construction, longitudinal patient representation, multi-view drug representation and medication combination prediction. The architecture of DNMDR is demonstrated in Figure \ref{fig2}.

\textbf{Dynamic graph construction} This module is responsible for the generation of edge weights, with which to guide the extraction of implicit structural relationships in diverse medical events of a patient’s single visit. 

\textbf{Longitudinal patient representation} With the dynamic networks constructed based on sematic features and structural information in EHR data, the LSTM-based GATs are leveraged to capture both temporal dependency and structural association for comprehensive patient representation.

\textbf{Multi-view drug representation} Integrating internal view information in drug molecules, such as drug atoms, chemical bones to compose single molecule, and interactive view relationships in drug pairs, the multi-view EHR graph and DDI graph are innovated which contribute to the integral drug representation.

\textbf{Medication combination prediction} Benefiting from the innovative contributions from the longitudinal patient representation and multi-view drug representation, the accurate and safe medication combination is output for special patient with a multi-label classifier.

\subsection{Dynamic graph construction}
\label{subsec42}
To model the evolution of diseases over time in the patient’s sequential EHRs, the RNN-based models, such as GRU\cite{yang2021safedrug,shang2019gamenet} and LSTM\cite{yu2024aka}, have yielded encouraging results on the patient’s semantic representations. Specifically, they learn the diagnosis codes, procedure codes and medication codes in the patients’ historical visits respectively, and leverage a further concatenation operation to fulfill the patients' representations for downstream tasks. In the encoding process, RNN-based models focus on the temporal nature of EHR data, while disregard the structural relationships in different medical events, leading to poor structural information in the patients' representations. Considering the advantages of graph data structure in modeling pairwise interactions between nodes, several studies have paid attention to graph embedding learning for the patient representations. With static graphs constructed on EHR data, the existing graph-based patient representations ignore the dynamically evolving of nodes and edges, which do not reflect the medical events and their varying relationships over time. Hence, the dynamic networks are constructed in this paper, which response to reflect both the evolution of temporal dependencies in historical EHRs and the dynamic nature of structural relationships between diverse medical events in a single visit.

To simplify the description of dynamic network construction, we take the EHR data of a single patient in his $t$-th visit as an example. As shown in Figure \ref{fig1}, each snapshot of a dynamic network repnses to the diverse medical events and their relationships in a patient’s visit to the hospital. With no definite connection explicitly between two neighboring nodes in a snapshot, we investigate the relationships among diverse medical events in a patient’s visit with a data statistical method.

\subsubsection{The construction of a snapshot}
\begin{figure*}
\centering
\includegraphics[scale=0.6]{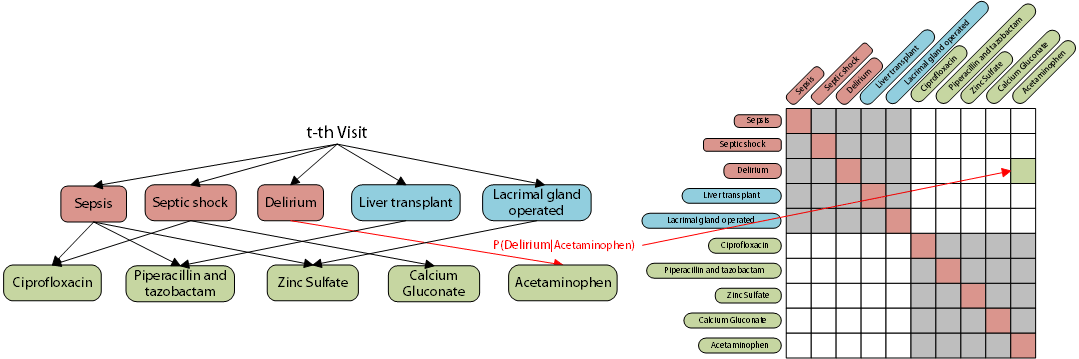}
\caption{Constructing a conditional probability matrix for the patient in a visit}
\label{fig3}
\begin{minipage}{\textwidth}
    \raggedright
    \textbf{*} The cells in gray represent the disallowed connections, and the red cells illuminate the guaranteed connections, and the white cells denote the allowed connections.
\end{minipage}
\end{figure*}

Considering the process of a physician in prescription making, each drug ordered mostly corresponds to the medical events encountered in the current visit. Given the EHR data in Figure \ref{fig3} as an example, the drug of Calcium Gluconate is ordered mostly according to the diagnosis of Septic shock, but not results from the procedure of Liver transplant. Moreover, in a patient’s EHR data, a drug prescription made is likely to comply with the connection between the medication codes and the diagnosis ones, while little is linked to the treatment codes. Therefore, to construct a snapshot for a patient’s dynamic network, we consider both the characteristic of EHR data and the conditional probabilities between features for structural relationships exploration.

Taking the practical clinical scenarios into account, we design a mask M to illuminate the structural relationships of diverse medical events with importance weights, which reflects the characteristic of EHR data and will be utilized during the attention generation step. Firstly, as to the connection with no practical significance for the MR task in this paper, it is forbidden in a snapshot and the mask M is set to negative infinity, such as the fact that a diagnostic node is not allowed to have any connection to the other diagnostic node or a procedure one. Secondly, for the nodes with connections to themselves, the mask M of the self-connected edges are assigned with positive infinity. Thirdly, since most of the nodes are allowed to have a connection to their neighboring nodes with a certain probability, the default value of zero is assigned for further calculation based on the conditional probabilities. The mask M of edges are initialized as follows.
{\small\begin{equation}M_{ij}=\begin{cases}+\infty&\text{a self-connection is encountered for i=j}\\0&\text{a connection is allowed with probability}\\-\infty&\text{a connection is of little significance in MR}\end{cases}\label{eq:equation2}\end{equation}} where $M_{ij}$ is the present weight for an edge connecting the neighboring nodes i and j.

To investigate the potential connections between two sets of features belonging to diverse medical events in a visit, we calculate the conditional probabilities between different nodes with data statistic method. Given the EHR data in Figure \ref{fig3} as an example, with no structural information explicitly available among the medical codes of Calcium Gluconate, Septic shock and Liver transplant, the conditional probabilities are calculated to investigate the reason why the drug of Calcium Gluconate is ordered. With computation on two sets of features in the three medical events respectively, the results illuminate that the probability $p(Calcium\, Gluconate \mid Septic\, shock)$ is dramatically larger than another one $p(Calcium\, Gluconate \mid Liver\, transplant)$. Therefore, we draw the conclusion that the connection between the former nodes are more likely exist than the one between the latter ones. Following this principle, we obtain the weights of the edges between heterogeneous nodes for a weighted snapshot of the patient’s EHR dynamic network. The formula for weight update is shown in Eq.\ref{eq:equation3}.

\begin{equation}{w}_{ij}^t(e_{ij})=P(i\mid j)=\frac{(n_{ij})}{(n_i)}\label{eq:equation3}\end{equation}
where $i$ and $j$ are heterogeneous nodes of edge $e_{ij}$, and ${w_{ij}}^t$ is the weight function to calculate the connection probability between nodes $i$ and $j$. Here, ${{n_{ij}}}$ is the number of co-occurred features in both medical nodes $i$ and $j$ of a patient’s visit, and ${{n_{i}}}$ is the number of features belonging to the medical events of node $i$.

With the conditional probabilities $p(m_{i}\mid d_{i})$ and $p(m_{i}\mid p_{i})$ calculated, the connection probabilities in weight matrix $P$ (shown in Figure \ref{fig3}) is updated and a normalization operation on each row is further performed. Since the mask $M$ and the conditional probability matrix $P$ are of the same size, the hidden structural information for weighted edge is calculated.
\begin{equation}e_{ij}^{ t}:=\hat{A}_{t}=\sigma(P+M_{ij})\label{eq:equation4}\end{equation}
where $\sigma$ is the normalization function.

\subsubsection{Building a dynamic network}
With original EHR of a patient’s $t$-th visit input, the clinical events with diverse medical codes $V_t=[d_t,p_t,m_t]$ are available, and the mapping function ${w_{ij}}^t({e_{ij}})$ is utilized to calculate connection probability for adjacency matrix ${\widehat A_t}$. In this way, the weighted edges with hidden structure information are obtained. Benefiting from a set of triads $(V_{t},E_{t},W_{t})$ achieved for a visit, we develop the $t$-th snapshot $G^{t}= (V_{t},E_{t},W_{t})$ for a patient’s dynamic network with the EHR data of $t$-th visit. In this way, a series of snapshots with discrete time for each visit are constructed based on the patient’s historical EHRs. Subsequently, these snapshots are combined sequentially to form a dynamic network that reflects the temporal evolution of a patient's health conditions with various medical events, in which each subgraph denotes the diverse medical activities in a patient’s visit to the hospital. Based on the weighted dynamic networks, DNMDR is capable to learn patient representations with both semantic information in nodes and structural relationships in edges.

\subsection{Longitudinal patient representation}
\label{subsec43}
\subsubsection{Node embeddings for a snapshot}
For a visit with diverse medical codes, three embedding tables $E_d\in\mathbb{R}^{|D|\times\mathrm{dim}}$, $E_p\in\mathbb{R}^{|P|\times\mathrm{dim}}$ and $E_m\in\mathbb{R}^{|M|\times\mathrm{dim}}$ are introduced for diagnosis, procedure and medication codes, respectively. Here, each row is an embedding vectors, and $dim$ is the dimension of the embedding space.

The original EHR data input are commonly multi-hot code vectors of diagnoses ${d_t}$, procedures ${p_t}$ and medications ${m_{t-1}}$, which imply the current health status of patient in the $t$-th visit. With the vector-matrix dot product, the multi-hot code vectors are projected into corresponding embedding vectors of diagnoses $d_t^e$, procedures $p_t^e$ and medications $m_{t-1}^e$, respectively.
\begin{equation}d_t^e=d_t\cdot E_d\label{eq:equation5}\end{equation}
\begin{equation}p_t^e=p_t\cdot E_p\label{eq:equation6}\end{equation}
\begin{equation}m_{t-1}^e=m_{t-1}\cdot E_m\label{eq:equation7}\end{equation}
where $d_t^e$, $p_t^e$ and $m_{t-1}^e$ are the embeddings of diagnosis, procedure and medication, respectively.

\subsubsection{Parameter evolution in snapshots}
Networks are ubiquitous in modeling the semantic features and structural relationships in real life applications. And most existing studies focus on static networks, in which the structure of vertices is fixed. Generally, the combination of GNNs with cyclic architecture is approved to generate the fixed node embeddings for static network representations. However, the sequence of a patient’s historical visits is developed with a dynamically evolving network including a series of snapshots in temporal order. Therefore, the node embeddings should be updated to reflect the temporal evolution of the structural relationships according to the variation of the patient’s health condition over time. Hence, the dynamicity and temporality in EHRs pose significant challenges in learning patient representations with the visit data in EHRs.

\begin{figure}
\centering
\includegraphics[scale=0.58]{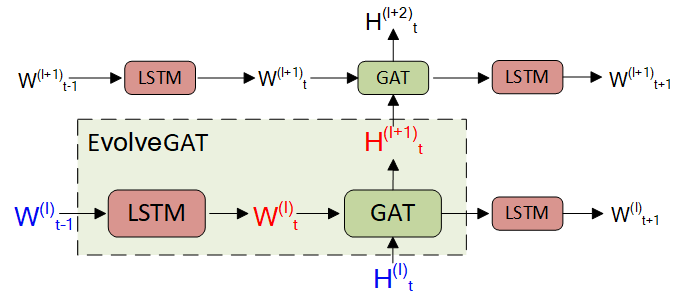}
\caption{The structure of a EvolveGAT}\label{fig4}
\begin{minipage}{0.48\textwidth}
    \raggedright
    \textbf{*} The blue illuminates the input, while the red denotes the output in the EvolveGAT unit.
\end{minipage}
\end{figure}

Benefiting from the present success of GNNs for static networks, we extend the GAT\cite{kipf2016semi} structure to dynamic setting, with an RNN-based architecture\cite{cho2014RNN} to learn the parameter evolution in snapshots. As shown in Figure \ref{fig4}, We propose a novel GAT evolution unit (EvolveGAT) . Specifically, for simplicity and effectiveness, we construct a GAT indexed by time for each visit firstly. Subsequently, the multi-layered GAT is leveraged to learn the structural relationships in diverse medical events within a snapshot of a visit. Furthermore, a Long Short-Term Memory(LSTM) architecture is introduced to evolve the GAT weights in snapshot sequence and update the parameters of the GATs in chronological order. 

In the process of learning a dynamic network for a patient’s integral representation, the key step lies in the weight updating of GATs with the evoluation of a patient’s temporal visits, which is beneficial to capture the dynamicity of structural relationships among diverse medical events in visits of each patient. To explore the evolution patterns of a dynamic network, the LSTM architecture is introduced which learns the dynamically updated hidden states of the snapshots in chronological order. The parameters in GATs is updated as follows:
\begin{equation}W_t^{(l)}=LSTM(W_{t-1}^{(l)})\label{eq:equation8}\end{equation}
where the initial weight $W_0^{l}$ is set to 1.

In DNMDR, some extensions are exerted on general LSTM architecture to learn the weights for GAT evolution. Firstly, the diverse medical codes input and the hidden states in GAT are extended from vectors to matrices. To obtain the input matrices, we simply set the column vectors side by side to form a matrix, and the same LSTM is deployed to process each column of the GAT weight matrix. Secondly, the dimension of the input column data is matched with the dimension of the hidden state. Finally, we perform the recurrent architecture with Eq.\ref{eq:equation9}.
\begin{equation}W_{t}^{(l)}=LSTM(W_{t-1}^{(l)}):=f(W_{t-1}^{(l)})\label{eq:equation9}\end{equation}
where $f(\cdot)$ is the calculation function.

In the DNMDR model, the multi-layered GAT for each visit is deployed which learns the structural information between diverse medical events within a snapshot. Taking the GAT for the t-th visit as an example. With the adjacency matrix $\hat{A}_{t}\in \mathbb{R}^{|V_t|\times |V_t|}$ and the node embedding matrix $H_{t}^{(l)}\in \mathbb{R}^{|V_t|\times dim}$ input, a weight matrix ${W_t^{(l)}}$ is leveraged to update the node embedding matrix for output, which is formulated as follows:
\begin{eqnarray}
H_{t}^{(l+1)}   &=& GAT(\hat{A}_{t},H_{t}^{(l)},W_{t}^{(l)})
\label{eq:equation11}
\end{eqnarray}
where $l$ is the layer index of GAT. With initial embedding matrix of node features $H_t^0=V_t$ input, GAT is capable to learn the high-level embeddings for network nodes, which belong to the snapshot of the patient’s $t$-th visit.

\subsubsection{Patient representions}
With each GAT to learn the network embeddings in a snapshot of a visit, DNMDR captures the semantic features and structural relationships in each snapshot and evolves the dynamic network with a series of snapshots in temporal order. At the same time, the feature vectors of each medical event are updated as follows:
\begin{equation}   
\begin{array}{c@{}r}  
\{H_{1:t-1}^d,H_{1:t-1}^p,H_{1:t-1}^m\} \\  
=EvolveGAT(\{d_{1:t-1}^e,p_{1:t-1}^e,m_{1:t-1}^e\},\hat{A}_{1:t-1})
\end{array} 
\label{eq:equation12}  
\end{equation}

With network embeddings obtained, the pooling layer is deployed to project medical codes of the same type in a visit into their individual embedding spaces.
\begin{equation}h_{1:t}^d=Pool(H_{1:t}^d)\label{eq:equation13}\end{equation}
\begin{equation}h_{1:t}^p=Pool(H_{1:t}^p)\label{eq:equation14}\end{equation}
\begin{equation}h^m_{1:t-1}=Pool(H_{1:t-1}^m)\label{eq:equation15}\end{equation}
where $t\in(1,T)$ denotes the $t$-th snapshot of a dynamic network corresponding to the patient’s $t$-th admission. Here, the contextual embeddings $h_t^d$,$h_t^p$ and $h_{t-1}^m$ are generated on basis of the diagnosis, procedure and medication codes respectively, each of which incorporates both the initial featural embeddings of nodes and the hidden structural information of edges with edge weights from the patients’ EHR dynamic network.

Subsequently, the GRUs are leveraged to learn the evolving of the snapshots and obtain the hidden diagnosis vector $h_t^{d'}$ and procedure vector $h_t^{p'}$, as well as the hidden historial medication vector $h_{t-1}^{m'}$, which are formulized as follows:
\begin{equation}h_t^{d'}=GRU_d\big(h^d_1,h^d_2,\cdots,h^d_t\big)\label{eq:equation16}\end{equation}
\begin{equation}h_t^{p'}=GRU_p\big(h^p_1,h^p_2,\cdots,h^p_t\big)\label{eq:equation17}\end{equation}
\begin{equation}h_{t-1}^{m'}=GRU_m\big(h^m_1,h^m_2,\cdots,h^m_{t-1}\big)\label{eq:equation18}\end{equation}

Combining the diagnosis embeddings with procedure ones, the longitudinal patient representation is generated by concatenation operation with a feedforward neural network.
\begin{equation}q_t=f([h_t^{d'},h_t^{p'}])\label{eq:equation19}\end{equation}
where $f(\cdot)$ is the transform function implemented with a single hidden layer fully connect neural network.

\subsection{Multi-view drug representation}
\label{subsec44}
\sloppy
On drug representation learning, prior approaches mainly focus on various fingerprints or drug profiles. These hand-crafted representations are limited by already known drug knowledge, while ignore the flexibility in discovering latent features beyond a physician’s experience. Currently, graph-based representation learning has emerged which extracts features from raw molecular graphs and captures potential structural relationships between drug pairs. Nevertheless, some researches lay emphasis on the atom and bond features in drug molecules, and other studies pay attention to the interaction patterns in drug pairs. As a consequence, there is little work to fully explore the drug molecule graphs from multiple level and various perspective. Therefore, DNMDR is designed to leverage the graph-based drug representation learning for prediction performance improvement in MR tasks.

To gain comprehensive drug representations for safe medication combination prediction, we simultaneously consider atom and bond features, drug molecule graphs, EHR and DDI graphs from multiple perspectives. First of all, the MPNN encoders are employed to capture atom and bond features of drugs based on structural similarity. Moreover, EHR and DDI graphs are constructed with the patients’ historical visits and known DDI knowledge respectively, in order to thoroughly capture drug co-occurrences and known DDIs in the EHRs. Additionally, an enriched knowledge base in the form of key-value pairs is leveraged to match the patient’s health status in current visit with the historical drug combinations, which offers the medical record-based retrieval process in the EHRs and simulates the efficient prescription making in clinical practice. As a consequence, the DNMDR model explores the drug combinations with both semantic information and structural relationships at atom-level, bone-level, molecule-level and drug-level, with the integration to achieve final low dimensional drug representations for medication combination prediction.

\subsubsection{Internal-view of drug molecules}
Different from the existing research on molecule-level drug analysis, we explore the drug molecular graph from the internal-view of drug molecules, investigating drug atoms and their bonds with both semantic features and structural function inside the drug molecules.

Firstly, a drug molecular graph $G_m = (a_i, A_m)$ is constructed, in which each atom is a node and each bond is an edge, indicating the atom-atom connectivity of the drugs, with the corresponding chemical bonds inside. Here, $a_i$ is the initial atom fingerprints from a learnable atom embedding table  $E_a\in\mathbb{R}^{|a_i|\times dim}$. And $A_m$ is the adjacency matrix illuminating the structural connection in node pairs.

Secondly, a message-passing neural network (MPNN) is leveraged to learn the drug molecular graph for node embedding vectors with real-value. To start with, each atom node in the drug molecular graph is initialize with its atom fingerprints $a_i$ , and each bond $e_{ij}\in A_m$ is initialize with its directly connected nodes $a_i$ and $a_j$ by their feature vectors. In the following message passing process, each atom is updated by aggregating the embedding vectors of neighboring nodes. Considering that the same type of chemical bonds in neighboring nodes have similar effects in structural information, the parameters in MPNN encoder are shared for drug molecules, and are set to the nodes corresponding to the same type of chemical bonds. The process is formulized as follows:
\begin{equation}
e_{a_i}^{(l)}=\sum\limits_{j\in \mathcal{N}(i)}{{{f}_{m}}}(h_{a_i}^{(l-1)},h_{a_j}^{(l-1)},{{w}_{ij}})
\label{eq:equation20}
\end{equation}
where ${{f}_{m}}\left( \cdot  \right)$ is a message passing function, $w_{ij}$ is a matrix of trainable parameters shared by the same type of chemical bond $e_{ij}$ at the $l$-th layer, $h_{a_i}^{(l-1)}$ is the hidden state of central atom $a_i$ at the $(l-1)$-th layer, $e_{a_i}^{(l-1)}$ is the embedding vector of central atom  $a_i$ computed by aggregating its neighboring atoms $\mathcal{N}(i)$ . And then, the hidden state is updated as follows:
\begin{equation}
h_{a_i}^{(l)}={{f}_{u}}(h_{a_i}^{(l-1)},e_{a_i}^{(l)})
\label{eq:equation21}
\end{equation}
where ${{f}_{u}}\left( \cdot  \right)$ is the update function. With the message passing for $L$ iterations, we obtain the final hidden states $h_{a_i}^{(L)}$ for all atoms.

Thirdly, with an aggregation operation on the hidden states, a fixed-size drug molecular embedding is available. To alleviate the overfitting issue, the pooling operation is exerted to calculate the average of all atom fingerprints.
\begin{equation}
{{h}_{m_i}}=Pooling(\{h_{a_j}^{(T)}\mid j=0,1,...,n\}) 
\label{eq:equation22}
\end{equation}
where $n$ is the number of atoms making up the drug molecule.

Finally, the similar MPNN operations with shared parameters are deployed on all the drug molecules, and the drug embeddings in internal-view are obtained for a drug memory $M_s \in \mathbb{R}^{|M| \times dim}$ , in which each row is the embedding vector of a specific drug.

\subsubsection{Interactive-view of drug molecules}
We explore drug co-occurrences and known DDIs in an interactive view on drug molecules, investigating EHR and DDI graphs with both semantic features and structural function at the molecule-level and drug-level.

Firstly, to capture drug co-occurrences within the EHRs, an adjacency matrix $A_{ehr}$ is employed to reflect all the drug pairs in each historical prescription. Moreover, another adjacency matrix $A_{ddi}$ is developed to illuminate all the known adverse DDIs should be avoided in recommended drug combinations. Combining the drug feature vectors in drug memory  $M_{s}$ with drug structural relationships in adjacency matrices, the EHR graph \(G_{ehr} = (M_{s}, A_{ehr})\) and the DDI graph \(G_{ddi} = (M_{s}, A_{ddi})\) are constructed in the interactive-view based on drug embeddings.

Secondly, we leverage two-layer GATs\cite{velickovic2017graph} to capture EHR information and DDI relationships for integral drug embeddings with both drug co-occurrences and adverse DDIs, respectively. Superior to the previous studies, we extend the GCN with an attention mechanism, and deploy the GAT model to weight neighboring nodes according to their importance to the current one, which is capable to deal with the potential structural problems and gains measurable relationships in drug pairs. As a result, the significant interactions are prioritized and the medication representations are refined delicately.

With the attention mechanism equipped, a GAT model executes the graph learning process by calculating the local network of drug nodes considering the importance of each DDI. Taking the DDI graph with $N$ nodes as an example. A linear transformation on each node of the feature vector is necessary to form a transformed feature space with $N$ nodes $h_{M}=\{h_{m_{1}},h_{m_{2}},...,h_{m_{i}},h_{m_{j}},...,h_{m_{N}}\} (1\leq\mathrm{i}\leq\mathrm{j}\leq\mathrm{N})$, in which each node $h_{m_i}$ contains a $dim$-dimensional feature vector with diverse features. The attention coefficient ${\alpha }_{ij}$ between drug nodes $m_i$ and its neighboring node $m_j$ is calculated as follows:
\begin{equation}
{{\alpha }_{ij}}=\frac{\exp (LeakyReLU({{a}^{T}}[W{{h}_{{{m}_{i}}}}\|W{{h}_{{{m}_{j}}}}]))}{\sum\limits_{k\in \mathcal{N}(i)}{\exp }(LeakyReLU({{a}^{T}}[W{{h}_{{{m}_{i}}}}\|W{{h}_{{{m}_{k}}}}]))} 
\label{eq:equation23}
\end{equation}
where $W$ is the weight matrix, $a^T$ is the transpose of the weight vector deployed on a single layer feed-forward neural network, $\|$ denotes the concatenation operation and ${{\cal N}(i)}$ is the neighboring set of the node $m_i$ in DDI graph. Here, ${h_{{m_i}}}$ and ${h_{{m_j}}}$ are the node embeddings and $LeakyReLU(\cdot)$ is the nonlinear activation function. Subsequently, the final output of node $m_i$ is linear combination of the normalized attention coefficient of each node in the set ${{\cal N}(i)}$ with its features. The process is written as
\begin{equation}
h_{{{m}_{i}}}^{'}=\sigma \left( \sum\limits_{j\in \mathcal{N}(i)}{{{\alpha }_{ij}}}W{{h}_{{{m}_{j}}}} \right) 
\label{eq:equation24}
\end{equation}
where $\sigma$ denotes a nonlinear activation function. In the similar manner, the node embedding table $E_{ddi} \in \mathbb{R}^{|M| \times dim}$ and $E_{ehr} \in \mathbb{R}^{|M| \times dim}$ are generated which are $dim$-dimensional feature vectors on DDI and EHR graphs respectively, indicating the forbidden adverse DDI relationships and acceptable drug co-occurrences in predicted medication combinations.

Finally, an integral drug representation $M_{ed}$ is obtained by integrating internal-view drug molecular information with interactive-view drug embeddings. Incorporating both semantic features and structural relationships on drug atom-level, chemical bond-level, drug molecule-level and drug pairs, the drug representation is formulized as follows:
\begin{equation}
M_{ed} = E_{ehr} + {WE}_{ddi}
\label{eq:equation25}
\end{equation}

\subsubsection{Temporal-view of drug prescription}

To learn historical drug combination and optimize the treatment in current visit, we design a dictionary structure based on key-value pairs $M_{kv}$, which offers a method for identifying the most similar patient representation over time and retrieving the appropriate weighted set of medications. Specifically, the patients $\begin{Bmatrix}q_1,\cdots,q_{t-1}\end{Bmatrix}$ and their corresponding multi-hot medication vectors $\begin{Bmatrix}h^{m'}_{1},\cdots,h^{m'}_{t-1}\end{Bmatrix}$ are provided in the form of key-value pairs, which are leveraged to locate the most matching patient with the proper medication combination by data retrieval. The incrementally inserted key-value pair with each visit is expressed as follows:
\begin{equation}M_{k\nu}^t=\left\{q_{i}:h^{m'}_{i}\right\}_1^{t-1}\label{eq:equation26}\end{equation}
\begin{equation}M_k^t=\begin{Bmatrix}q_1;q_2;\ldots;q_{t-1}\end{Bmatrix}\label{eq:equation27}\end{equation}
\begin{equation}M_{\nu}^{t}=\begin{Bmatrix}h^{m'}_{1};h^{m'}_{2};\ldots;h^{m'}_{t-1}\end{Bmatrix}\label{eq:equation28}\end{equation}
where $M_k^t, M_\nu^t\in \mathbb{R}^{\left|t-1\right|\times\dim}$ are key and value vectors at the $t$-th visit of a patient.

\subsection{Medication combination prediction}\label{subsec45}
With the drug embedding vectors in $M_s$, $M_{ed}$ and $M_{kv}^t$ achieved for the $t$-th visit, the longitudinal patient representation $q_t$ is input as a query to retrieve the most relevant drug combination, and the outputs $o^s_t$, $o^{ed}_t$ and $o^{kv}_t$ are achieved as follows:
\begin{equation}o^s_t=\sigma\left(M_sq_t\right)\label{eq:equation29}\end{equation}
\begin{equation}o^{ed}_t=M_{ed}^T\sigma\left(M_{ed}q_t\right)\label{eq:equation30}\end{equation}
\begin{equation}o^{kv}_t=M_{ed}^T{(M_{kv}^t)}^T\sigma\left(M_{kv}q_t\right)\label{eq:equation31}\end{equation}
where $\sigma$ is a activation function, $o^s_t\in \mathbb{R}^{\dim}$ is calculated with dot product between the MPNN-generated drug molecule embeddings $M_s$ and the longitudinal patient representation $q_t$. Here, $o^{ed}_t\in \mathbb{R}^{\dim}$ is obtained by combining drug co-occurrence and adverse DDI information. Furthermore, $o^{kv}_t\in \mathbb{R}^{\dim}$ is achieved with the context attention on the most matching patient representation in the historical visits.

Finally, with the comprehensive patient representation and the integral drug representation fed, the output vector for medication recommendation is calculated as follows:
\begin{equation}\hat{y}_{t}=\sigma(q_t,o^s_t,o^{ed}_t,o^{kv}_t)\label{eq:equation32}\end{equation}
where $\sigma$ is a activation function.

\subsection{Model training}
\label{subsec46}
In model training, a combination of three losses are leveraged for end-to-end medication recommendation. With the binary cross-entropy loss, multi-label hinge loss and drug-drug interaction loss calculated, the weighted summation is available to generate the final loss function.

Binary cross-entropy loss $\mathcal{L}_{bce}$ measures the discrepancy between the predicted labels and the true ones. For a patient in his $t$-th visit, the loss $\mathcal{L}_{bce}$ is calculated as follows:
\begin{equation}\mathcal{L}_{bce}=-\sum_{k=1}^{|M|}\biggl[y_t^{(k)}\log(\hat{y}_t^{(k)})+(1-y_t^{(k)})\log(1-\hat{y}_t^{(k)})\biggr]\label{eq:equation33}\end{equation}
where $|M|$ is the number of medications, $y_t^{(k)}$ is the ground truth label of medication $k$ at the $t$-th visit, and $\hat{y}_t^{(k)}$ is the predicted probability of the medication $k$ at the $t$-th visit.

The multi-label hinge loss $\mathcal{L}_{multi}$ is adopted to ensure that the score of the correct class is higher than that of the incorrect ones by a margin, which is defined as follows:
\begin{equation}\mathcal{L}_{multi}=\frac{1}{\left|M\right|}\sum_{i,j:m_i^t=1,m_j^t=0}\max\left(0,1-(\hat{y}_t^{(i)}-\hat{y}_t^{(j)})\right)\label{eq:equation34}\end{equation}

The drug-drug interaction (DDI) loss $\mathcal{L}_{ddi}$ penalizes the model for drug pairs with known adverse DDIs. With the DDI adjacency matrix $A_{ddi}$ given, $\mathcal{L}_{ddi}$ is calculated as follows:
\begin{equation}\mathcal{L}_{ddi}=\sum_{i=1}^{|M|}\sum_{j=1}^{|M|}\Big(A_{ddi}^{ij}\cdot\hat{y}_{t}^{(i)}\cdot\hat{y}_{t}^{(j)}\Big)\label{eq:equation35}\end{equation}

The final loss function is a weighted summation of the above three losses:
\begin{equation}\mathcal{L}=\alpha\mathcal{L}_{\mathrm{bce}}+\beta\mathcal{L}_{\mathrm{multi}}+\gamma\mathcal{L}_{\mathrm{ddi}}\label{eq:equation36}\end{equation}
where $\alpha$, $\beta$ and $\gamma$ are the given hyper parameters to balance the prediction loss and the DDI loss.

The pseudocode for one training epoch of proposed DNMDR is shown in Algorithm 1:
\begin{algorithm}[H]
\caption{One Training Epoch of DNMDR}
\begin{algorithmic}[1]
\Input Preprocessed training dataset $\tau$, hyperparameters;
\State Initialize parameters: $E_d, E_p, E_m, E_a, W_i$;
\State Obtain the EHR graph $G_{ehr}$, the DDI graph $G_{ddi}$;
\For {patient $j := 1$ to $|\tau|$}
    \State Select $j$-th patient's EHR data, $X_j$;
    %\State Obtain the drug memory $M_s$;
    
    \State// Dynamic Network Construction
    
    \State Build dynamic network for $j$-th patient by snapshot combination in temporal order $G_j=\left\{G_j^1,\dots, G_j^{|\tau|}\right\}$ in Eq.(\ref{eq:equation2}--\ref{eq:equation4}); 
    
    \For {visit $t := 1$ to $|X_j|$}
        \State Generate node emebddings for $t$-th snapshot with embedding vectors $d_t^e$, $p_t^e$, $m_{t-1}^e$ in Eq.(\ref{eq:equation5}--\ref{eq:equation7});
        
        \State Obtain the node embedding with EvolveGAT update on $H_{1:t-1}^d$,$H_{1:t-1}^p$ and $H_{1:t-1}^m$ in Eq.(\ref{eq:equation8}--\ref{eq:equation12});
        
        \State Obtain the node embedding with GRU update on $h_{t}^d$,$h_{t}^p$ and $h_{t-1}^m$ in Eq.(\ref{eq:equation13}--\ref{eq:equation18});
        
        \State //Patient Representation
        
        \State Generate patient representation $q_t$ in Eq. \ref{eq:equation19} 
        
        \State Generate the drug molecular embedding $M_s$ in internal-view with Eq.(\ref{eq:equation20}--\ref{eq:equation22});
        
        \State Generate the drug embedding $M_{ed}$ in interactive-view of drug molecules with Eq.(\ref{eq:equation23}--\ref{eq:equation25});
        
        \State Generate the dictionary of drug embeddings based on key-value pairs $M_{kv}^{t}$ with Eq.(\ref{eq:equation26}--\ref{eq:equation28});

        \State// Multi-view Drug Representation
        
        \State Generate multiple-view drug representation $o^s_t$, $o^{ed}_t$ and $o^{kv}_t$ with Eq.(\ref{eq:equation29}--\ref{eq:equation31}); 

        \State// Medication Combination Prediction
        
        \State Generate recommended drug combination $\hat{y_t}$ with Eq.\ref{eq:equation32} 

    \EndFor
    \State Accumulate $\mathcal{L}_{bce}, \mathcal{L}_{multi}, \mathcal{L}_{ddi}$ in Eq.(\ref{eq:equation33}--\ref{eq:equation36});
\EndFor
\end{algorithmic}
\end{algorithm}

\section{Experiments}
\label{sec5}
We conduct extensive experiments to evaluate the performance of DNMDR on openly available real-world datasets, with the commonly used metrics.

\subsection{Datasets}
\label{subsec51}
In the experiments, three datasets on EHRs, DDIs and drug molecules are utilized for performance evaluation on DNMDR, with comparison to the baseline models. 

The MIMIC-III dataset\cite{johnson2016mimic} is a publicly available EHR dataset widely used in the medical domain, which contains clinical data of 46,520 patients admitted to the Beth Israel Deaconess Medical Center in Boston, Massachusetts, between 2001 and 2012. There are a variety of attributes in the MIMIC-III dataset, such as demographics, vital signs, laboratory test results, medications, diagnoses, procedures, and so on. In the MIMICIII dataset, Each patient is represented with a unique identifier and is associated with one or more hospital admissions and ICU stays. In the experiments, the diagnosis, procedure and medication information is selected for performance evaluation. Specifically, the ICD-9 coding system is adopted for diagnoses and procedures, while the NDC (National Drug Code) system is offered for medications. The dataset statistics after preprocessing are summarized in Table \ref{tbl2}.
\begin{table}
\caption{The MIMIC statistics after preprocessing}\label{tbl2}
\begin{tabular}{@{}L|C@{}}
\toprule
Items & Number \\ 
\midrule
\# of visit / \# of patients & 1,4995 / 6,350 \\ 
diag. / prod. / med. space size & 1958 / 1430 / 131 \\ 
avg. / max \# of visit & 2.37 / 29 \\ 
avg. / max \# of diagnoses per visit & 10.51 / 128 \\ 
avg. / max \# of procedures per visit & 3.84 / 50 \\ 
avg. / max \# of medications per visit & 11.44 / 65 \\ 
total \# of DDI pairs & 448 \\ 
total \# of substructures & 491 \\
\bottomrule
\end{tabular}
\end{table}

To extract DDI information, the TWOSIDES dataset is widely used, which consists of 645 drugs, 63473 drug pairs, and 1317 interactions in drug pairs. With DDIs and side effects in drug pairs reported by ATC (Anatomical Therapeutic Chemical) Three Level codes., we aim to improve the detection and prediction of adverse drug effects and interactions in MR tasks. Based on the TWOSIDES dataset, we build a DDI adjacency matrix for DDI graph construction, in which the top 40 severity types are gravely concerned and should be protected from the possible adverse consequences.

For the EHR graph construction, we transform the medications of NDC drug codes in MIMIC dataset to ATC ones and build a drug co-occurrence adjacency matrix, thereby facilitates the score calculation on DDIs for data integration in drug representation generation.

With the drug molecules in SMILES strings collected from drugbank, we build the drug molecule graph, in which the atoms are nodes and the chemical bonds are edges connecting neighboring nodes. We learn the graph representations for comprehensive drug molecular embeddings.

\subsection{Experimental settings}
\label{subsec52}
In the experiments, we split the MIMIC dataset into training, validation, and test sets with a ratio about 4:1:1. The hyperparameter tuning is performed on the validation set, resulting in the optimal ones for model training. In model training, the Adam optimizer with a learning rate of 0.0001 is adopted, and a dropout rate of 0.4 is applied to the input embedding layer for avoiding overfitting. All the experiments are conducted with PyTorch and trained on the NVIDIA RTX 4090 GPU.

\subsection{Evaluation Metrics}
\label{subsec53}
In the evaluation experiments, five widely used evaluation metrics are adopted: Precision Recall Area Under Curve (PRAUC), Jaccard similarity score (Jaccard), F1 score, DDI rate and the number of drugs. The higher values of PRAUC, F1 score and Jaccard indicate the better model performance, whereas the lower values of DDI rate and the number of drugs illuminate the better result of medication recommendation.

\textbf{F1 Score} is the harmonic mean of precision and recall, offering a balance between the two. It is calculated as
\begin{equation}Precision_t\quad=\frac{\mid Y_t\cap\hat{Y_t}\mid}{\mid Y_t\mid}\label{eq:equation37}\end{equation}
\begin{equation}Recall_t=\frac{\mid Y_t\cap\hat{Y}_t\mid}{\mid\hat{Y}_t\mid}\label{eq:equation38}\end{equation}
\begin{equation}F1=\frac{1}{T}\sum_{t=1}^T\frac{2\times Precision_t\times Recall_t}{Precision_t+Recall_t}\label{eq:equation39}\end{equation}

\textbf{Precision Recall Area Under Curve (PRAUC)} treats each drug to be recommended with a probability and evaluates the precision and the recall at different thresholds. The formula for PRAUC calculation is shown as follows:
\begin{equation}\Delta Recall(k)_t=Recall(k)_t-Recall(k-1)_t\label{eq:equation40}\end{equation}
\begin{equation}PRAUC=\frac{1}{T}\sum_{t=1}^{T}\sum_{k=1}^{|M|}Precision(k)_{t}\Delta Recall(k)_{t}\label{eq:equation41}\end{equation}
where $Precision(k)_{t}$ is the precision at cut-off $k$ in the ordered retrieval list for visit $t$, and $Recall(k)_{t}$ is the gap in recall from rank $k-1$ to $k$.

\textbf{Jaccard Similarity Score} measures the overlapping between predicted drug set and the true ones. It is calculated as a ratio between the intersection and the union of them two.
\begin{equation}Jaccard=\frac{1}{T}\sum_{t=1}^T\frac{\mid Y_t\cap\hat{Y_t}\mid}{\mid Y_t\cup\hat{Y_t}\mid}\label{eq:equation42}\end{equation}
where $Y_t$ is the set of true drugs and $\hat{Y_t}$ is the set of predicted drugs for visit $t$.

\textbf{DDI Rate} measures the safety of drug combinations recommended by calculating the percentage of DDIs within the recommended results. The DDI rate is defined as follows:
\begin{equation}DDI\,rate=\frac{\sum_{t=1}^T\sum_{i,j}\mathbf{1}(m_{t,i},m_{t,j}\in m_t\mathrm{~and~}A_{ddi}^{i,j}=1)}{\sum_{t=1}^T|m_t|}\label{eq:equation43}\end{equation}
where $A_{ddi}$ is the drug matrix, $m_t$ is the drug set in the $t$-th visit, and \textbf{1} is the indicator function denotes the true result complying with the given rules.

\textbf{Number of Drugs} captures the average number of drugs recommended in each visit. In MR tasks, it is important for the model not to recommend an excessive number of drugs. Otherwise, it may lead to polypharmacy and increased risk of adverse effects. The number of drugs is calculated with the average of all drugs prescribed in each visit.
\begin{equation}Number=\frac{1}{T}\sum_{t=1}^{T}\left|m_{t}\right|\label{eq:equation44}\end{equation}

\subsection{Baseline models}
\label{subsec54}
For performance evaluation, we compare the DNMDR with several representative baseline models in MR tasks. To ensure a fair comparison, all the models are run on the preprocessed MIMIC-III dataset with the same experimental setup. The baselines are outlined as follows:

\begin{itemize}
\item \textbf{Logistic Regression (LR)}\cite{luaces2012binary} is an instance-based classifier, which employs L2 regularization to prevent overfitting and is widely used for binary classification tasks.

\item \textbf{Ensemble Classifier Chain (ECC)}\cite{read2009classifier} is a well-known multi-label classification approach with multi-hot vectors of diagnoses and procedures input. ECC models the interdependencies among labels with an ensembles of classifier chains.

\item \textbf{RETAIN}\cite{choi2016retain} employs a reverse time attention mechanism combined with an RNN to enhance the interpretability of prediction results by focusing on significant past visits.

\item \textbf{LEAP}\cite{zhang2017leap} is an LSTM-based generative model to recommend drugs sequentially, considering the patient's diagnostic history in making medication decisions.

\item \textbf{GAMENet}\cite{shang2019gamenet} stores historical drug records in memory and utilizes memory-augmented neural networks as references, ensuring safe and effective drug combination recommendations.

\item \textbf{MICRON}\cite{yang2021change} features a residual-based inference mechanism for sequential updating. It adjusts the patient's health condition with new clinical data while maintains previous drug combinations unchanged.

\item \textbf{SafeDrug}\cite{yang2021safedrug} leverages the drug molecule and substructure graphs to guide medication relationship modeling, aiming to provide safe and effective drug recommendations.

\item \textbf{COGNet}\cite{wu2022conditional} incorporates a novel copy-or-predict mechanism, allowing to copy medications from previous recommendations or to predict new ones for the current medication set.

\item \textbf{AKA-SafeMed}\cite{yu2024aka} is a attention-based representation model that employs a knowledge augmentation strategy for drug recommendations.
\end{itemize}

\begin{table*}[htbp]
\centering
\caption{Performance Comparison on MIMIC-III}
\begin{tabular}{l|ccccc}
\hline

\hline
Model & Jaccard & F1 Score & PRAUC & DDI Rate & Avg. \# of Drugs \\ \hline \hline
LR & 0.4949 $\pm $ 0.0023 & 0.6518 $\pm $ 0.0022 & 0.7559 $\pm $ 0.0020 & 0.06799 $\pm $ 0.0010 & 16.55 $\pm $ 0.0488 \\ 
ECC & 0.4807 $\pm $ 0.0022 & 0.6368 $\pm $ 0.0021 & 0.7560 $\pm $ 0.0021 & 0.07913 $\pm $ 0.0008 & \textbf{15.81 $\pm $ 0.1999} \\ \hline
RETAIN & 0.4834 $\pm $ 0.0017 & 0.6448 $\pm $ 0.0015 & 0.7598 $\pm $ 0.0006 & 0.08409 $\pm $ 0.0016 & 18.39 $\pm $ 0.4539 \\ 
LEAP & 0.4492 $\pm $ 0.0020 & 0.6113 $\pm $ 0.0019 & 0.6524 $\pm $ 0.0020 & 0.06789 $\pm $ 0.0010 & 18.82 $\pm $ 0.0497 \\ 
GAMENet & 0.5063 $\pm $ 0.0017 & 0.6624 $\pm $ 0.0015 & 0.7647 $\pm $ 0.0017 & 0.08329 $\pm $ 0.0014 & 25.94 $\pm $ 0.2010 \\ 
MICRON & 0.5174 $\pm $ 0.0013 & 0.6721 $\pm $ 0.0011 & 0.7739 $\pm $ 0.0003 & 0.06210 $\pm $ 0.0007 & 17.92 $\pm $ 0.1322 \\ 
SafeDrug & 0.5161 $\pm $ 0.0016 & 0.6720 $\pm $ 0.0015 & 0.7671 $\pm $ 0.0008 & 0.06165 $\pm $ 0.0008 & 19.6 $\pm $ 0.3136 \\ 
COGNet & 0.5141 $\pm $ 0.0028 & 0.6677 $\pm $ 0.0029 & 0.7413 $\pm $ 0.0089 & 0.08317 $\pm $ 0.0025 & 25.06 $\pm $ 0.4202 \\
AKA-SafeMed & 0.5164 $\pm $ 0.0021 & 0.6725 $\pm $ 0.0018 & 0.7709 $\pm $ 0.0016 & 0.06171 $\pm $ 0.0010 & 20.84 $\pm $ 0.4219 \\ \hline
\textbf{DNMDR} & \textbf{0.5241 $\pm $ 0.0011} & \textbf{0.6787 $\pm $ 0.0009} & \textbf{0.7749 $\pm $ 0.0008} & \textbf{0.06058 $\pm $ 0.0007} & 19.53 $\pm $ 0.2664 \\ \hline

\hline
\end{tabular}
\label{tb:Table3}
\end{table*}
\begin{figure*}[htbp]
    \centering
    \subfigure[Jaccard \& Epochs]{
        \includegraphics[width=0.3\textwidth]{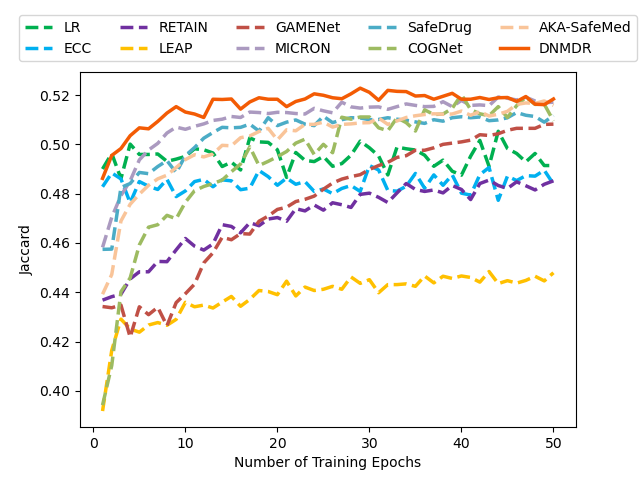}
    }
    \subfigure[F1 \& Epochs]{
        \includegraphics[width=0.3\textwidth]{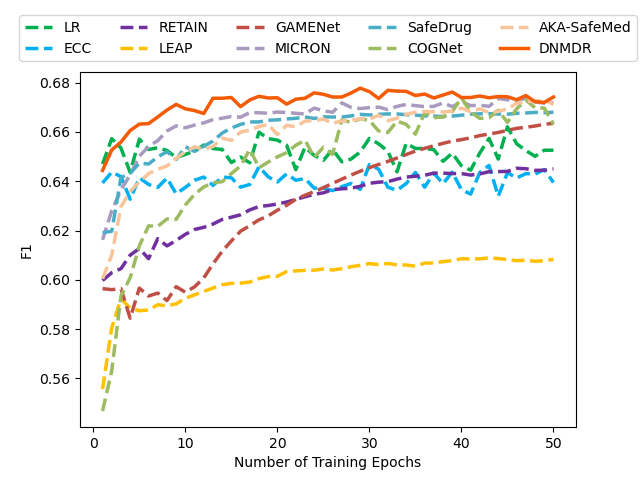}
    }
    \subfigure[PRAUC \& Epochs]{
        \includegraphics[width=0.3\textwidth]{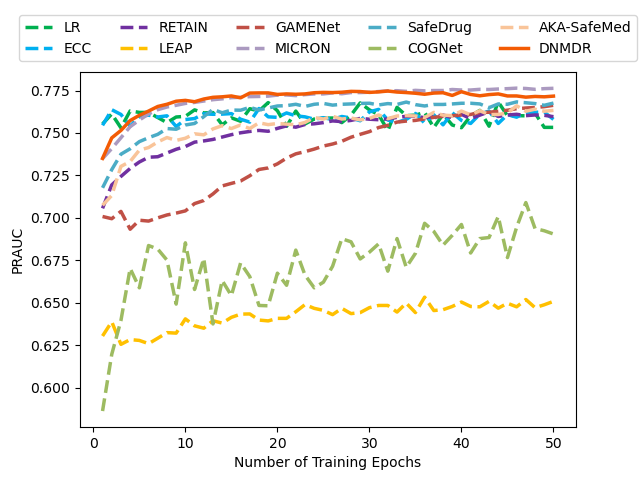}
    }
    \subfigure[DDI Rate \& Epochs]{
        \includegraphics[width=0.3\textwidth]{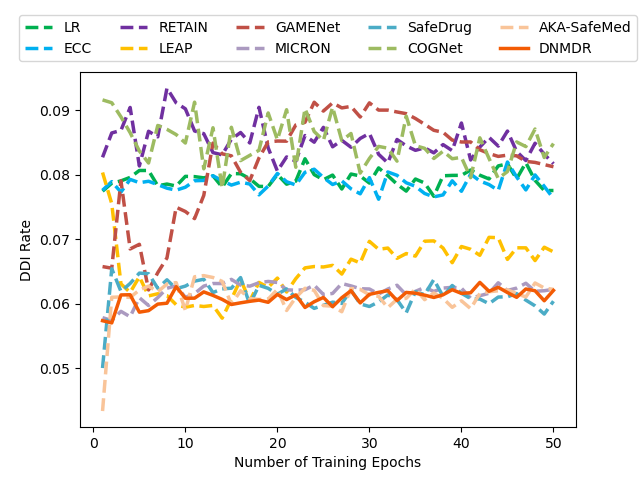}
    }
    \subfigure[Avg. \# of Drugs \& Epochs]{
        \includegraphics[width=0.3\textwidth]{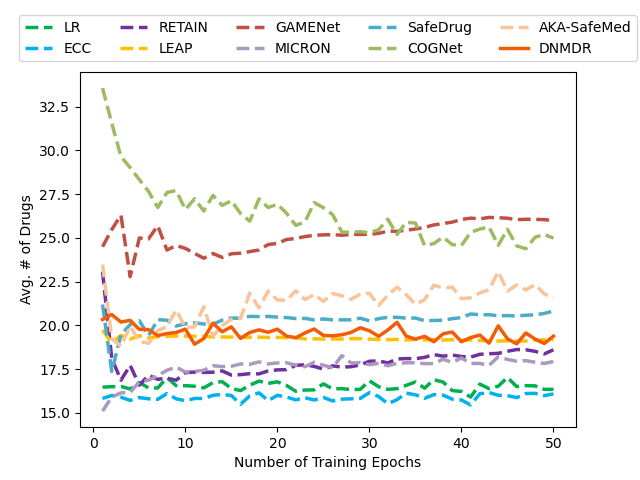}
    }
    \subfigure[Best epoch \& Epochs]{
        \includegraphics[width=0.3\textwidth]{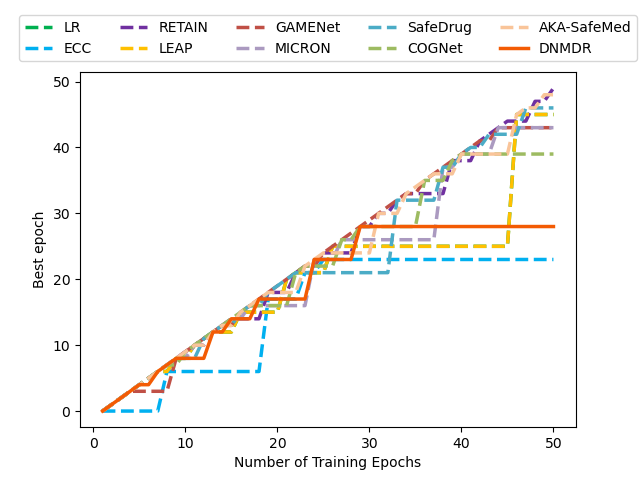}
    }
    \caption{Performance variation with the epochs in comparison experiments.}
    \label{fig:Figure5}
\end{figure*}

\subsection{Experimental results}
\label{subsec55}
The experimental results on five commonly used metrics are shown in Table \ref{tb:Table3}, and the performance variations on different models with the training epochs are shown in Figure \ref{fig:Figure5}. Comparing to eight generally used baseline models, the DNMDR model demonstrates outstanding performance in comparison results of Jaccard, F1 score, PRAUC and Number of Drugs, while shows a competitive DDI rate, which outperforms all the baselines. Some experimental results are summarized as follows.

Firstly, at the initial stage of model training, the evaluation indicators of Jaccard, F1, PRAUC and DDI rates are gradually raised as the training epochs increase. As the training goes on, DNMDR shows consistent improvements and runs into a stable state with less epochs, compared with its competitive models. Consequently, DNMDR outperforms all the baseline models in terms of Jaccard, PRAUC, DDI rate and F1 score with less epochs experienced.

Secondly, the instance-based models such as LR and ECC show lower performance than the other ones, due to the fact that they focus on data in current visit while ignore the roles of historical healthcare information. The temporal models like RETAIN perform better by incorporating the patients’ visit history, though its high DDI rate is notable due to the large number of drug combinations recommended.

Thirdly, longitudinal approaches such as GAMENet, MICRON, and SafeDrug show improved performance by integrating additional information on drugs. Specifically, GAMENet leverages historical drug records and graph information, MICRON retains previous drug combinations with a residual-based method, and SafeDrug incorporates drug molecular structures to reduce DDI rates. COGNet utilizes the similarity of historical visits for medication generation, achieving an efficient recommender. AKA-SafeMed employs the self-attention mechanism to learn the weights of medical vectors from a patient’s historical visits. However, the inherent DDI rate in historical prescriptions (around 0.08) leads to a higher DDI rate in COGNet and results in much more medications in recommended drug list. Therefore, there is a promising task to balance the accuracy and the safety in drug recommendation.

Finally, Our DNMDR model achieves the best performance on metrics of Jaccard, F1, PRAUC and DDI rate, demonstrating its superior accuracy and safety in MR tasks. Notably, DNMDR  maintains a lower DDI rate compared to most of the baselines, highlighting its effectiveness in minimizing adverse DDIs. The success is mostly benefited from the dynamic network construction and multi-view drug information in DNMDR, and results from its accurate capturing the patients’ structural relationships in EHR data. Additionally, the architectural advantage of DNMDR lies in its effective learning semantic features and structural relationships in patients’ visit sequences, providing a robust solution for MR tasks. Generally, DNMDR excels in both accuracy and safety in comparison experiments.

\begin{table*}[htbp]
\centering
\caption{Ablation Study for Different Modules of DNMDR on MIMIC-III.}
\begin{tabular}{l|ccccc}
\hline

\hline
Model & Jaccard & F1 Score & PRAUC & DDI Rate & Avg. \# of Drugs \\ \hline \hline
DNMDR\_prob & 0.5210 $\pm$ 0.0016 & 0.6761 $\pm$ 0.0015 & 0.7733 $\pm$ 0.0009 & 0.06063 $\pm$ 0.0006 & 19.81 $\pm$ 0.3315 \\ 
DNMDR\_lstm & 0.5201 $\pm$ 0.0012 & 0.6752 $\pm$ 0.0010 & 0.7718 $\pm$ 0.0013 & 0.06103 $\pm$ 0.0008 & 19.76 $\pm$ 0.2727 \\ 
DNMDR\_internal & 0.5210 $\pm$ 0.0014 & 0.6764 $\pm$ 0.0012 & 0.7738 $\pm$ 0.0009 & 0.06039 $\pm$ 0.0005 & 19.80 $\pm$ 0.2965 \\ 
DNMDR\_interactive & 0.5178 $\pm$ 0.0013 & 0.6734 $\pm$ 0.0012 & 0.7706 $\pm$ 0.0007 & 0.06379 $\pm$ 0.0006 & 19.93 $\pm$ 0.3517 \\ 
DNMDR\_dynn & 0.5218 $\pm$ 0.0019 & 0.6768 $\pm$ 0.0017 & 0.7716 $\pm$ 0.0008 & \textbf{0.05955 $\pm$ 0.0009} & 20.03 $\pm$ 0.3999 \\ \hline
\textbf{DNMDR} & \textbf{0.5241 $\pm$ 0.0011} & \textbf{0.6787 $\pm$ 0.0009} & \textbf{0.7749 $\pm$ 0.0008} & 0.06058 $\pm$ 0.0007 & \textbf{19.53 $\pm$ 0.2664} \\ \hline

\hline
\end{tabular}
\label{tb:Table4}
\end{table*}
\begin{figure*}[htbp]
    \centering
    \subfigure[Jaccard \& Epochs]{
        \includegraphics[width=0.3\textwidth]{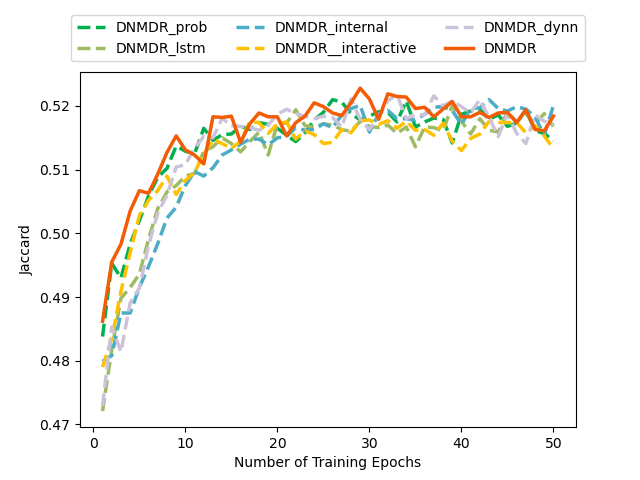}
    }
    \subfigure[F1 \& Epochs]{
        \includegraphics[width=0.3\textwidth]{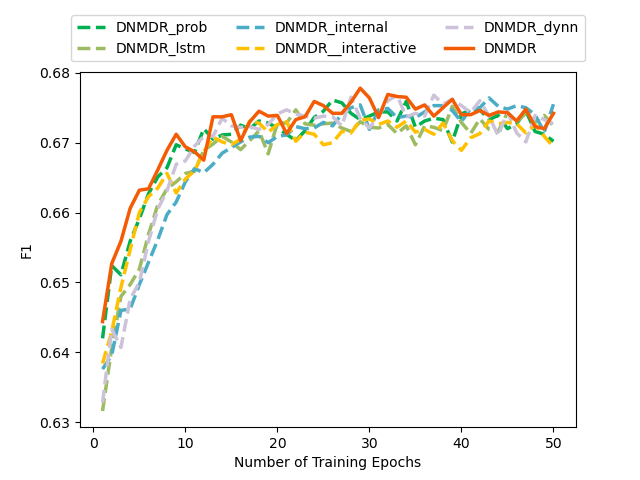}
    }
    \subfigure[PRAUC \& Epochs]{
        \includegraphics[width=0.3\textwidth]{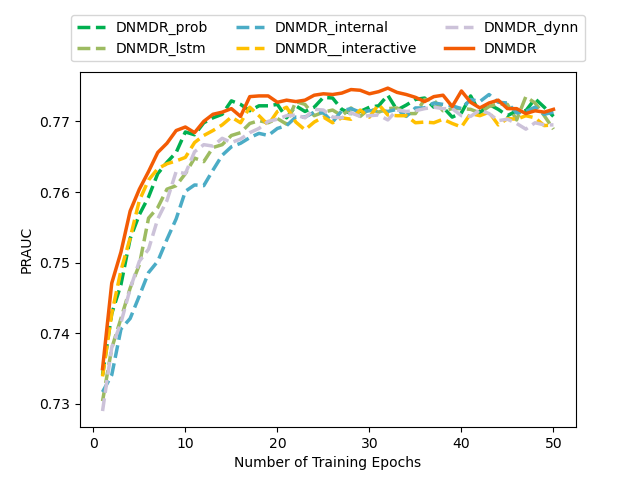}
    }
    \subfigure[DDI Rate \& Epochs]{
        \includegraphics[width=0.3\textwidth]{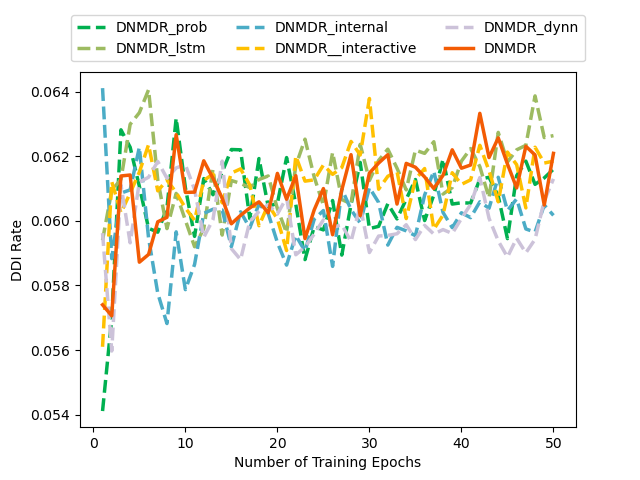}
    }
    \subfigure[Avg. \# of Drugs \& Epochs]{
        \includegraphics[width=0.3\textwidth]{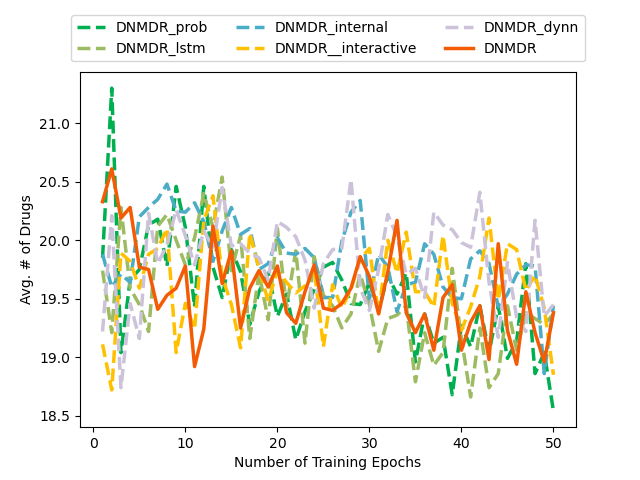}
    }
    \subfigure[Best epoch \& Epochs]{
        \includegraphics[width=0.3\textwidth]{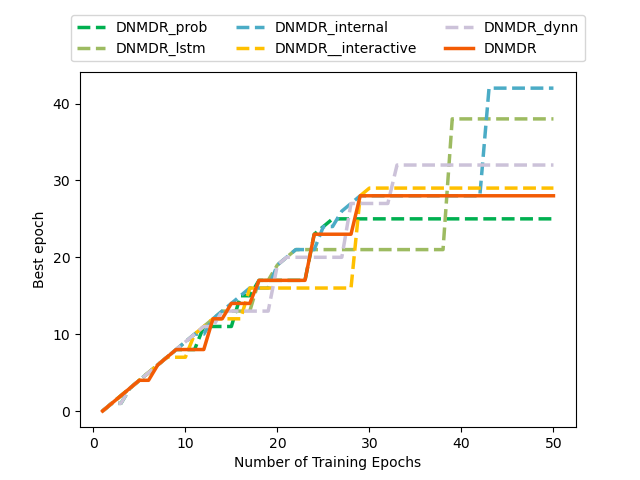}
    }
    \caption{Performance variation of different ablation models with the epochs}
    \label{fig:Figure6}
\end{figure*}

\subsection{Ablation Study}\label{subsec56}
To better investigate the contribution of each component for the overall performance of 
DNMDR model, we conduct an ablation study with the following variants derived from DNMDR:
\begin{itemize}
\item DNMDR\_prob: We remove the calculation process of conditional probability, treating all the medical nodes of a visit in a fully connected state. This means that the implicit instruct information $\hat{A}_t$ is an all-ones matrix in Eq. \ref{eq:equation4}.
	
\item DNMDR\_lstm: We remove the LSTM model for GAT parameter evolving in patient representation learning based on the dynamic networks.

\item DNMDR\_internal: We remove the internal-view information in drug molecule graph learning.

\item DNMDR\_interactive: We remove the interactive-view information in drug molecule graph learning.

\item DNMDR\_dynn: We remove the dynamic network construction, and learn patient representation with sequential EHR data.
\end{itemize}
\begin{table*}[htbp]
\centering
\caption{The performance of DNMDR in different dimensions.}
\begin{tabular}{c|ccccc}
\hline

\hline
Dimension & Jaccard & F1 Score & PRAUC & DDI Rate & Avg. \# of Drugs\\ \hline
64 & 0.5241 $\pm$ 0.0011 & 0.6787 $\pm$ 0.0009 & 0.7749 $\pm$ 0.0008 & 0.06058 $\pm$ 0.0007 & 19.53 $\pm$ 0.2664 \\ 
128 & 0.5283 $\pm$ 0.0019 & 0.6810 $\pm$ 0.0016 & 0.7795 $\pm$ 0.0014 & 0.06020 $\pm$ 0.0008 & \textbf{19.11 $\pm$ 0.3215} \\ 
\textbf{256} & \textbf{0.5309 $\pm$ 0.0014} & \textbf{0.6847 $\pm$ 0.0012} & \textbf{0.7816 $\pm$ 0.0007} & \textbf{0.05980 $\pm$ 0.0005} & 20.34 $\pm$ 0.2728 \\ 
512 & 0.5284 $\pm$ 0.0026 & 0.6817 $\pm$ 0.0022 & 0.7779 $\pm$ 0.0019 & 0.06002 $\pm$ 0.0008 & 20.13 $\pm$ 0.3524 \\ \hline

\hline
\end{tabular}
\label{tb:Table5}
\end{table*}

The experimental results with different variants of DNMDR are demonstrated in Table \ref{tb:Table4}. It reveals that no model in the table wins the other ones in all the five metrics. Specifically, the proposed DNMDR model achieves the best performance on accuracy target in MR tasks, while DNMDR\_dynn manifests the strongest safety in ablation study. The results illuminate the structural advantage of dynamic network construction for comprehensive patient representations in MR tasks. Additively, it is notable that the DNMDR\_dynn model achieves a much lower DDI rate than DNMDR. it may be result from the certain amount of DDIs reported in the original MIMIC dataset. More specifically, in the historical prescriptions made for the patients, there are DDIs with the ratio of approximate 8\% in the drug prescriptions recorded in EHR data of the MIMIC-III dataset\cite{yang2021safedrug}.

In ablation experiments, the model training  processes on DNMDR and its variants are briefly explored in Figure \ref{fig:Figure6}, which illuminates the consistent improvements of all the models.

First, all of the six models exhibit gradually rise on metrics related to accuracy and reach the relative stable state in performance. The training processes indicate the good convergence characteristic with steady downward trend on loss, which is crucial to ensure the stability and reliability of each model. 

Second, the poor performance of DNMDR\_prob in all the five metrics exhibits the advantage of dynamic graph learning in patient representation generation. based on the condition probabilities calculated and the connection mask assigned, the weights are generated with different importance on dynamical edges, indicating structural relationships in diverse medical events. Combining semantic features and structural information, DNMDR is beneficial to learn comprehensive patient representation, which further facilitates the downstream drug representation and medication prediction tasks.

Third, with ignored LSTM structure, the DNMDR\_lstm model manifests slightly inferior performance, which may due to the absence in learning the evolving parameters in sequential GATs. the experiments on DNMDR\_lstm show that the evolution of layer parameters have contributions to the final recommendation results.

Finally, both DNMDR\_internal and DNMDR\_interact suffer extreme performance degradation on all the five metrics, and the ablation study reveals the necessity of multi-view and hierarchical-level drug molecule graph learning in integral drug representation leading to effective and safe drug recommendation. The former captures drug molecule structure with different importance in drug atom-level and chemical bond-level, while the latter weights the edges in EHR and DDI graphs to identify the drug co-occurrences and adverse DDIs in molecule-level and drug-level. Therefore, the combination of drug knowledge and potential DDIs are beneficial to achieve safe drug representations.

In summary, our DNMDR model outperforms all the other variants on the Jaccard, F1 and PRAUC evaluation metrics. On the one hand, the dynamic networks contribute the comprehensive patient representations. On the other hand, the multi-view drug molecule exploration benefits the safe drug representation. Hence, the ablation study confirms that each component in DGMRM is essential and DNMDR is effective and safe to achieve final medication combination recommendation.

\subsection{Parametric Sensitivity}\label{subsec57}
\subsubsection{Effect of vector dimension}
To investigate the effect of varying vector dimensions on the performance of DNMDR , we set the dimensions of the medical embedding vectors derived from EHR data to 64, 128, 256, and 512, respectively. The experimental results with different vector dimensions are shown in Table \ref{tb:Table5}. When the dimension of medical embedding vectors is 256, the DNMDR model performs best in the experiments. The increasing or decreasing on embedding vector dimension results in reduced performance for DNMDR. Moreover, a suitable increase in vector dimension achieves enhanced performance of DNMDR, which is likely due to the richer and more valuable clinical information encapsulated within the vectors. However, as the vector dimension becomes too large, the DNMDR model encounters performance degradation, which may be resulted from overfitting caused by the high-dimensional vectors. Therefore, an appropriate vector dimension is essential for improving the generalization capability of the DNMDR model.
\begin{table*}
\centering
\caption{The example patient's record for case study}
\begin{tabular}{cm{4cm}m{4cm}m{4cm}}
\hline

\hline
 & Diagnoses & Procedures & Medications \\
\hline
visit1 & 2749, 2449, 4019, \textcolor{blue}{56400}, 33819, 2930, V4986, 1985, V103, V1052, 28522, 1970, 73313, 3383, 3363, E8700, 99709,\,\, 34931 & 8848, 8104, 3979, 8105, 8163, 8451, 0359, 8099, 8321, 8844 & N02B, A01A, A02B, A06A, B05C, A12A, A12C, M01A, N01A, N02A, A02A, B01A, C10A, H03A, J01D, N03A, N05A, A04A, N05B, C09C, H04A \\
\hline
visit2 & 5119, 2749, 78552, 99592, 4019, 0389, \textcolor{blue}{56400}, 5990, 2930, 99813, V4986, 27652, 1985, V103, 0413, V1052, 28522, V454,\,\, 1970,\,\,\,\, 04149 & 3491 & N02B, A01A, A06A, B05C, A12A, A12C, A07A, C03C, A12B, N02A, J01M, B01A, H03A, J01D, N05A, A04A, N05B,\,\, R01A,\,\, J01E,\,\, M03B \\
\hline
visit3 & 2749, \textcolor{red}{99859}, E8788, 30000, \textcolor{red}{V1254}, 5121, 1985, V103, V1052, 99832, 28522, \textcolor{Green}{1970}, V153,\,\,\,\,\,\, 1972,\,\, 9986,\,\, 51189 & \textcolor{red}{3404}, \textcolor{red}{8344} & N02B, A02B, \textcolor{blue}{A06A}, A12C, A07A, M01A, C03C, \textcolor{orange}{A12B}, \textcolor{orange}{C02D}, N02A, \textcolor{red}{B01A}, C10A, H03A,\,\, A04A,\,\, N05B \\
\hline

\hline
\end{tabular}
\label{tb:Table6}
\end{table*}

\begin{figure}[htbp]
  \centering
    \includegraphics[width=0.48\textwidth]{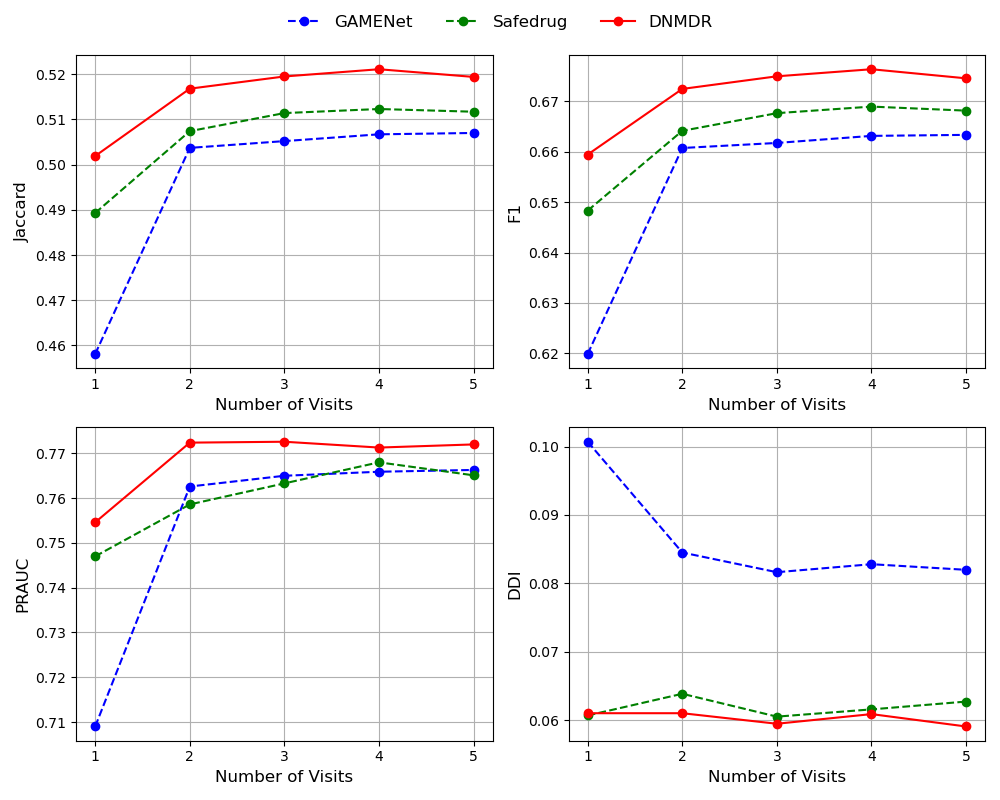}
    \caption{ The effect of number of visits for various models.}\label{fig7}
\end{figure}
\subsubsection{Effect of number of visits}
In order to explore how the number of visits affects the performance of DNMDR, we deploy  multiple longitudinal MR methods on the EHRs with different number of visits. Considering  limited patients with more than four visits in the MIMIC dataset, we take the EHR data for patients with the last five visits for parametric sensitivity analysis. the results depicted in Figure \ref{fig7} demonstrate that the DNMDR model consistently outperforms GAMENet and Safedrug on MIMIC-III dataset. Furthermore, with the increasing of visiting numbers of the same patient, the DNMDR model experiences first performance improvement and then degradation, which indicates that the EHR data in recently historical visits do have a positive effect on the current recommendation, while the patient’s historical visits in the far past may mislead the current treatment with noise data unrelated to the present diseases. Therefore, we draw the conclusion that the DNMDR model is superior to its competitors at effectively modeling the patients’ representations and enhancing safety in drug combinations for recommendation.

\subsection{Case Study}\label{subsec58}

To illustrate the effect of DNMDR, we select a patient from the test dataset for a detailed analysis. As shown in Table \ref{tb:Table6}, There are diverse medical events of ICD codes and multiple medications of ATC codes in the patient’s three visits. With the diagnosis and procedure codes in the three visits and the medication codes in the first two visits input, we constrct the patient’s dynamic network. Briefly, two historical visits are leveraged to build the two snapshots, in which the diverse medical events are heterogeneous nodes and the structural relationships are dynamic edges with weights. The proposed DNMDR model is deployed for medication combination prediction, and the outputs are demonstrated in Table \ref{tb:Table7}.

\begin{figure}
  \centering
    \includegraphics[width=0.48\textwidth]{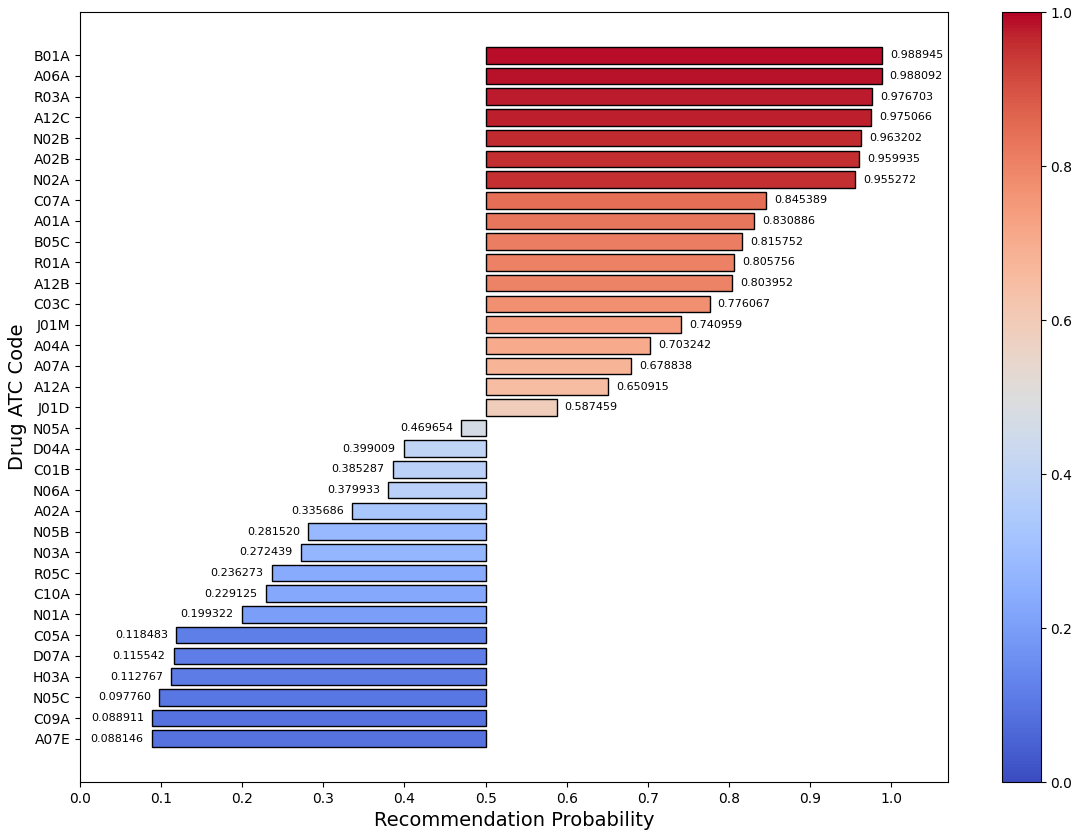}
    \caption{recommended medications with probability for interpretation}\label{fig8}
\end{figure}
\begin{table*}
\centering
\caption{Comparison of model recommended drug coding with the original data set}
\begin{tabular}{m{7cm}m{7cm}}
\hline

\hline
Prescribed medication encoding & Recommendation medication encoding \\
\hline
B01A, A06A, H03A, A12C, N02B, A02B, N02A, C10A, M01A, A12B, C03C, A04A, A07A, N05B, C02D   & B01A(0.98895), A06A(0.98809), R03A(0.97670), A12C(0.97507), N02B(0.96322), A02B(0.95994), N02A(0.95527), C07A(0.84539), A01A(0.83089), B05C(0.81575), R01A(0.80576), A12B(0.80395), C03C(0.77607), J01M(0.74096), A04A(0.70324), A07A(0.67884),\,\, A12A(0.65091),\,\,\,\, J01D(0.58746) \\
\hline

\hline
\end{tabular}
\label{tb:Table7}
\end{table*}
\begin{figure*}[htbp]
    \centering
    \subfigure[The heat map of the first visit in the patient’s dynamic network]{
        \includegraphics[width=0.48\textwidth]{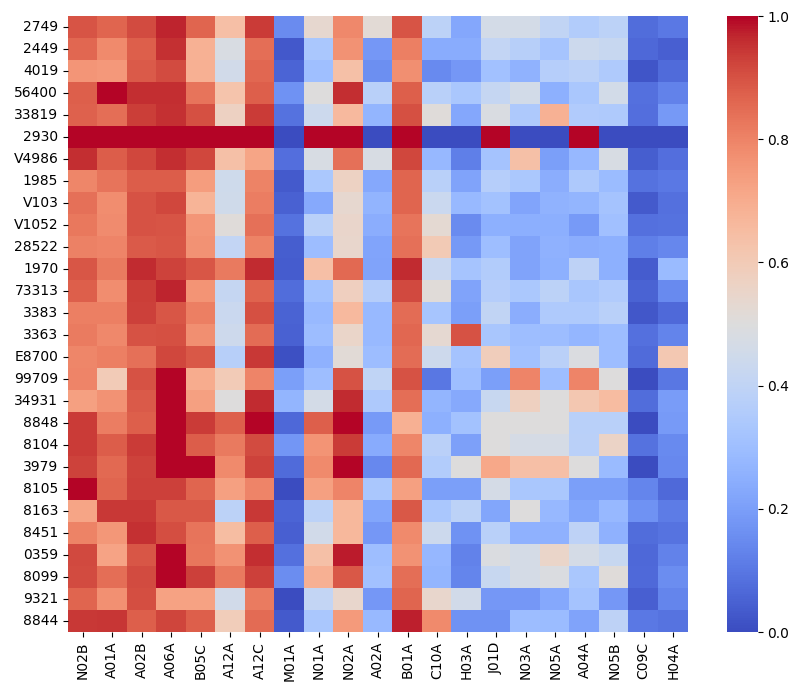}
    }
    \subfigure[The heat map of the second visit in the patient’s dynamic network]{
        \includegraphics[width=0.48\textwidth]{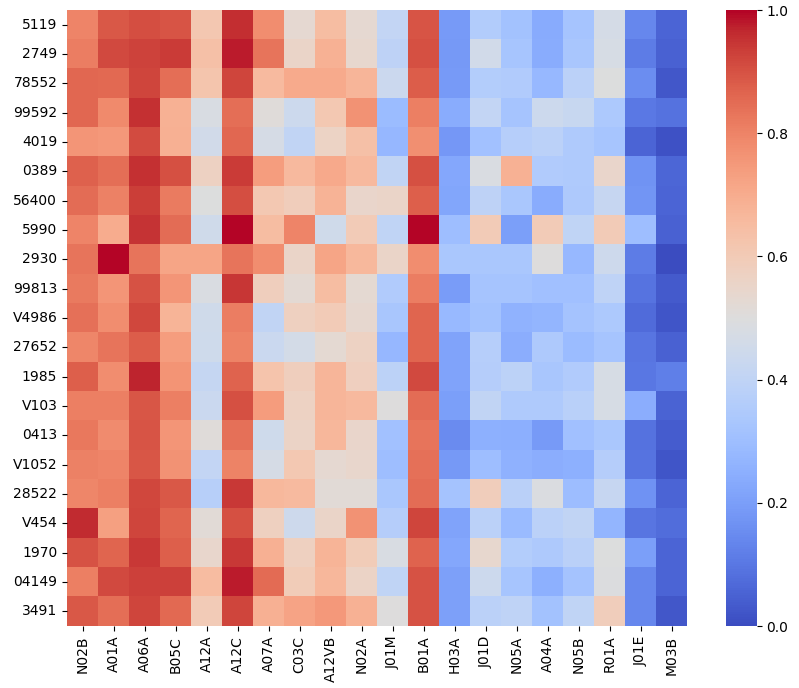}
    }
    \caption{Heat map of attention between medical nodes in dynamic network}
    \label{fig:Figure9}
\end{figure*}

Firstly, we take the recommended medication identical to the drug in the current prescription as an example. As shown in Figure \ref{fig8}, the medication B01A (Displayed in red in Table 6) gains the highest recommendation rate of 0.98895, which is an anticoagulant to prevent the formation of blood clots by interfering with certain steps in the blood clotting process. We detect the original EHR data and find that the medication B01A is available in the doctor's prescription, which is consistent with the clinical facts in MIMIC-III dataset. To explore the prescription evidence of included drug B01A, we analyze the diagnoses and procedures in the patient’s third visit. As a possible evidence, the code 998.59 (Displayed in red in Table 6) is a post-operative complication, indicating the patient’s increased risk of venous thromboembolism (VTE) after surgery. Such a risk is particularly high in patients with long-term immobility or high-risk clinical activities. As another possible evidence, we know that encoding V1254(Displayed in red in Table 6) represents the patient’s personal history of stroke, indicating a probable vascular disease or a tendency to clot. For this reason, the drug B01A may be recommended to prevent a repeat clot event on the patient. Still another evidence, encoding 3404(Displayed in red in Table 6) implies the procedures to insert, remove, or replace a pacemaker, while encoding 8344(Displayed in red in Table 6) is cardiac catheterization. As we know, both cardiac surgery and interventional procedures can increase the risk of blood clots on a patient. Therefore, based on the possible evidences mentioned above, it is likely reasonable and necessary for DNMDR to recommend the drug B01A in the patient’s third visit.

Secondly, an example is taken to investigate the properly recommended medication for the evolving chronic disease. Drug A06A (Displayed in blue in Table 6) is recommended and prescribed as a type of laxative drug, while there is no diagnosis related to the symptom of constipation in the patient’s third visit. Analysis on the previous two visits, we notice that the diagnostic code 56400 (Displayed in blue in Table 6) is constipation, which is recurrent to imply a chronic constipation in the patient’s medical history. As a chronic problem for long-term care, it is reasonable for a doctor to prescribe A06A for a chronic constipation prevention before definitive diagnosis in each visit. With drug A06A recommended, the DNMDR model shows its learning ability on complex interaction and evolution in the patient’s historical records. The attention heat map shown in Figure \ref{fig:Figure9} reveals the snapshots of the first two visits in the patient’s dynamic network. The strong correlation of drug A06A in the two visits further highlights the effectiveness of DNMDR in learning evolving sequential EHR data.

Additionally, we notice that some recommended drugs are not really prescribed by the doctors in the clinical scenarios. We take drug R03A (Displayed in green in Table 6) as an example, and analyze the possible reasons according to the patient's current diagnoses and procedures. with the drug knowledge available, we know that the drug R03A is a $\beta$-2 adrenergic agonist to relax bronchial smooth muscle and relieve bronchospasm by activating $\beta$-2 receptors on the cells. It is generally used for asthma, chronic obstructive pulmonary disease (COPD) and to relieve bronchospasm.  Considering that the diagnostic code 1970 (Displayed in green in Table 6) represents a second malignant new organism: lung (metastatic lung cancer), we infer that drug R03A ($\beta$-2 adrenergic agonist) has certain therapeutic effect on the symptom of dyspnea and bronchospasm caused by 197.0 (Displayed in green in Table 6) (metastatic lung cancer). In this sense, we suggest that recommended drug R03A by DNMDR is reasonable in the clinical scenarios.

It is worth mentioning that a drug combination of C02D (Displayed in orange in Table 6) (peripheral alpha-blocker) and A12B (Displayed in orange in Table 6) (potassium supplement) is available in the patient’s third visit of the original MIMIC-III dataset, while the simultaneously taking these two medications can cause side effects, such as difficult breathing, gastric ulcers, neuropathy, atopic dermatitis, heart failure and so on. Furthermore, the drug pairs of C02D (Displayed in orange in Table 6) and A12B have adverse side effects on respiratory, digestive, nervous, immune, cardiovascular and other systems. With this in mind, the DNMDR model offers C07A (Displayed in orange in Table 6) (peripheral alpha-blocker) instead of C02D (beta-blocker). With both drugs to treat hypertension, the DNMDR recommender alleviates DDIs in a drug combination.

\section{Conclusion}\label{sec6}
To enhance the safety and efficacy in medication recommendation, the DNMDR model is proposed in this paper. With dynamic network constructed, DNMDR is capable to integrate temporal evolution and structural relationships for comprehensive patient representations. Benefiting from the multi-view drug molecule graph learning, DNMDR generates safe drug representations for the downstream medication combination presentation. Experimental results on the MIMIC-III dataset demonstrate that DNMDR outperforms the state-of-the-art baseline models, with superior performance in Jaccard, F1, PRAUC metrics and low DDI rate. These improvements highlight the DNMDR model in providing accuracy and safety in MR tasks.

In the future research, we will further investigate the structural relationships in EHRs, and explore personalized medication recommendation based on the knowledge supporting from medical knowledge bases. With the powerful learning ability of Large Language Model (LLM), we direct our energies to personalized precise medication recommendation with controllable DDIs.

\section*{Acknowledgments}
\sloppy
This work is supported by the Natural Science Foundation of Shandong Province, China (No. ZR2021MF118, No. ZR2020LZH008, No. ZR2022LZH003), the Key R\&D Program of Shandong Province, China (No.2021CXGC010506, No.2021SFGC0104) and the Na tional Natural Science Foundation of China (No. 62101311, No. 62072290),Youth Science Foundation Project of Shandong Province (No. ZR2022QF022), Postgraduate Quality Education and Teaching Resources Project of Shandong Province (No. SDYKC2022053, No. SDYAL2022060), and Jinan “20 new colleges and universities” Funded Project (No. 202228110).
% Use the style of numbering in square brackets.
% If nothing is used, default style will be taken.
%\begin{enumerate}[a)]
%\item 
%\item 
%\item 
%\end{enumerate}  

%\begin{itemize}
%\item 
%\item 
%\item 
%\end{itemize}  

%\begin{description}
%\item[]
%\item[] 
%\item[] 
%\end{description}  

%\clearpage %%Remove this from your manuscript

% Figure
%\begin{figure}%[]
%  \centering
%    \includegraphics{}
%    \caption{}\label{fig1}
%\end{figure}

%\begin{table}%[]
%\caption{}\label{tbl1}
%\begin{tabular*}{\tblwidth}{@{}LL@{}}
%\toprule
%  &  \\ % Table header row
%\midrule
% & \\
% & \\
% & \\
% & \\
%\bottomrule
%\end{tabular*}
%\end{table}

% Uncomment and use as the case may be
%\begin{theorem} 
%\end{theorem}

% Uncomment and use as the case may be
%\begin{lemma} 
%\end{lemma}


%\section{}\label{}


\bibliographystyle{cas-model2-names}

\bibliography{refs}



\end{document}